\documentclass[sigconf]{acmart}
\AtBeginDocument{%
  }
\copyrightyear{2026}
\acmYear{2026}
\setcopyright{cc}
\setcctype{by}
\acmConference[WWW '26] {Proceedings of the ACM Web Conference 2026}{April 13--17, 2026}{Dubai, United Arab Emirates.}
\acmBooktitle{Proceedings of the ACM Web Conference 2026 (WWW '26), April 13--17, 2026, Dubai, United Arab Emirates}
\acmISBN{979-8-4007-2307-0/2026/04}
\acmDOI{10.1145/3774904.3792494}
\usepackage{amsmath}
\usepackage{xspace}
\usepackage{tikz}         % 核心绘图包
\usepackage{graphicx}     % 图像插入支持
\usepackage{pifont}
\usepackage{amsfonts}
\usepackage{array}
\usepackage{multirow}
\usepackage{bm}

\usepackage{balance}

\usepackage[table,dvipsnames]{xcolor} % 支持表格着色并提供更多颜色名（如 Teal, Purple）
\usepackage{colortbl}                 % （可选）扩展表格颜色命令（\cellcolor, \rowcolor）
\usepackage{booktabs}   

\usepackage{makecell}                  % 提供 \Xhline, \Xcline{a-b}{width} 等命令
\usepackage[table,dvipsnames]{xcolor}  % 可选：表格着色时用（提供更多颜色名）

\newcommand{\our}{\textsf{Prism}\xspace}
\newcommand{\ours}{\textsf{Prism-SFT}\xspace}
\newcommand{\ourd}{\textsf{Prism-DPO}\xspace}
\newcommand{\itiu}{\textsf{ITIU}\xspace}

\newcommand{\mi}{\textsf{Mistral-Interact}\xspace}

\newcommand{\ir}{\textsf{IR}\xspace}
\newcommand{\collab}{\textsf{\textsc{CollabLLM}}\xspace}
\newcommand{\bpo}{\textsf{BPO}\xspace}

\newcommand{\inth}{\texttt{IN3}\xspace}
\newcommand{\tin}{\texttt{TIN}\xspace}
\newcommand{\abp}{\texttt{ABP}\xspace}
\newcommand{\cid}{\texttt{CID}\xspace}

\newcommand{\myparagraph}[1]{\par\addvspace{1ex}\noindent\textbf{#1.}\hspace{1em}}

\newcommand{\down}[1]{$_{\color{BlueGreen}\downarrow #1}$}
\newcommand{\up}[1]{$_{\color{RedOrange}\uparrow #1}$}

\newcommand{\circlednum}[1]{\ding{\numexpr171+#1}}

\newcolumntype{L}[1]{>{\raggedright\let\newline\\\arraybackslash\hspace{0pt}}m{#1}}
\newcolumntype{C}[1]{>{\centering\let\newline\\\arraybackslash\hspace{0pt}}m{#1}}
\newcolumntype{R}[1]{>{\raggedleft\let\newline\\\arraybackslash\hspace{0pt}}m{#1}}
\begin{document}

% \title{\our: Towards Lowering User Cognitive Load in LLMs \\ via Complex Intent Understanding}

\title[Towards Lowering User Cognitive Load in LLMs via Complex Intent Understanding]{\texorpdfstring{\our: Towards Lowering User Cognitive Load in LLMs \\ via Complex Intent Understanding}{Prism: Towards Lowering User Cognitive Load in LLMs via Complex Intent Understanding}}

\author{Zenghua Liao}
\orcid{0009-0002-3155-2353}
\affiliation{%
  \department{National Key Laboratory of Big Data and Decision}
  \institution{National University of Defense Technology}
  \city{Changsha}
  \state{Hunan}
  \country{China}
}
\email{liaozenghua18@nudt.edu.cn}

\author{Jinzhi Liao}
\orcid{0000-0002-2898-6559}
\authornotemark[1]
\affiliation{%
  \department{National Key Laboratory of Big Data and Decision}
  \institution{National University of Defense Technology}
  \city{Changsha}
  \state{Hunan}
  \country{China}
}
\email{liaojinzhi12@nudt.edu.cn}

\author{Xiang Zhao}
\orcid{0000-0001-6339-0219}
\authornote{Corresponding authors}
\affiliation{%
  \department{National Key Laboratory of Big Data and Decision}
  \institution{National University of Defense Technology}
  \city{Changsha}
  \state{Hunan}
  \country{China}
}
\email{xiangzhao@nudt.edu.cn}

\renewcommand{\shortauthors}{Liao et al.}

\begin{abstract}

Large Language Models are rapidly emerging as web-native interfaces to social platforms.   
On the social web, users frequently have ambiguous and dynamic goals, making complex intent understanding—rather than single-turn execution—the cornerstone of effective human-LLM collaboration. 
Existing approaches attempt to clarify user intents through sequential or parallel questioning, yet they fall short of addressing the core challenge: modeling the logical dependencies among clarification questions.
Inspired by the Cognitive Load Theory, we propose \textbf{\our}, a novel framework for complex intent understanding  that enables logically coherent and efficient intent clarification.
\our comprises four tailored modules: 
a complex intent decomposition module, which decomposes user intents into smaller, well-structured elements and identifies logical dependencies among them; 
a logical clarification generation module, which organizes clarification questions based on these dependencies to ensure  coherent, low-friction interactions; 
an intent-aware reward module, which evaluates the quality of clarification trajectories via an intent-aware reward function and leverages Monte Carlo Sample to simulate user-LLM interactions for large-scale, high-quality training data generation; and 
a self-evolved intent tuning module, which iteratively refines the LLM's logical clarification capability through data-driven feedback and optimization.
\our consistently outperforms existing approaches across clarification interactions, intent execution, and cognitive load benchmarks.
It achieves state-of-the-art logical consistency, reduces logical conflicts to 11.5\%, increases user satisfaction by 14.4\%, and decreases task completion time by 34.8\%.
All data and code are released\footnote{~\url{https://github.com/liaozenghua/Prism}}.

% \vspace{-5pt}
\end{abstract}





\maketitle

\begin{tikzpicture}[remember picture,overlay,shift={(current page.north west)}]
\node[anchor=north west,xshift=2.0cm,yshift=-2.5cm]{\scalebox{0.5}{\includegraphics[width=2cm]{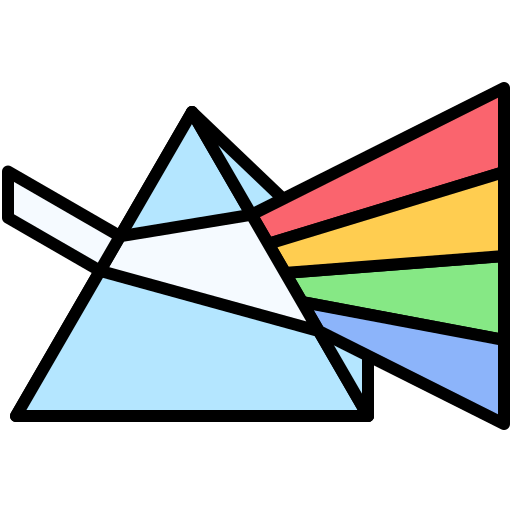}}};
\end{tikzpicture}

% \myparagraph{Relevance to the Web} 1324

\vspace{-25pt}
\section{Introduction}

% \todo{The full logic should be reorganized as:
% User intent is important -> previous studies tried -> they ignore complex intent -> they cannot address and what’s the reason -> challenge -> theory inspires us -> technical details}

% Modern 

The web has become the primary medium of human–LLM interaction, with LLMs assistants increasingly embedded within social platforms to help users discover, create, and manage information~\cite{DBLP:conf/www/Bouleimen25,DBLP:conf/www/ZhangSDL24}.
Modern LLMs excel at generating high-quality single-turn responses when given with well-specified input~\cite{DBLP:journals/corr/abs-2303-08774,DBLP:conf/icml/WuGP0LDC0L025}.
% ~\cite{DBLP:journals/corr/abs-2303-08774,DBLP:journals/corr/abs-2407-21783,DBLP:journals/corr/abs-2403-08295,DBLP:conf/icml/WuGP0LDC0L025}.
However, in web-mediated contexts, users usually not comprehensively understanding their intents, tend to initiate conversations with only vague descriptions~\cite{taylor:1968}. 
This \textit{gap} forces users to iteratively refine requests through successive clarification to fully meet their demands.
The effectiveness of this clarification process determines user experience and task success~\cite{guidelines, johnny,understand_user_experience,dissatisfaction}, thereby revealing the significance of \textit{understanding user intent}.
% ~\cite{guidelines, johnny,understand_user_experience,dissatisfaction,DBLP:conf/icml/WuGP0LDC0L025}

% Large language models (LLMs) excel at generating high-quality single-turn responses when given with well-specified input~\cite{DBLP:journals/corr/abs-2303-08774,DBLP:journals/corr/abs-2407-21783,DBLP:journals/corr/abs-2403-08295,DBLP:conf/icml/WuGP0LDC0L025}.
% However, real-world users, usually not comprehensively understanding their intents, tend to initiate conversations with only a vague description~\cite{taylor:1968}. 
% This \textit{gap} forces users to iteratively refine requests through successive clarification to fully meet their demands.
% The effectiveness of this clarification process determines user experience and task success~\cite{guidelines, johnny,understand_user_experience,dissatisfaction,DBLP:conf/icml/WuGP0LDC0L025}, thereby revealing the significance of \textit{understanding user intent}.

\begin{figure}[!t]
\centering
  
\includegraphics[width=1.0\linewidth]{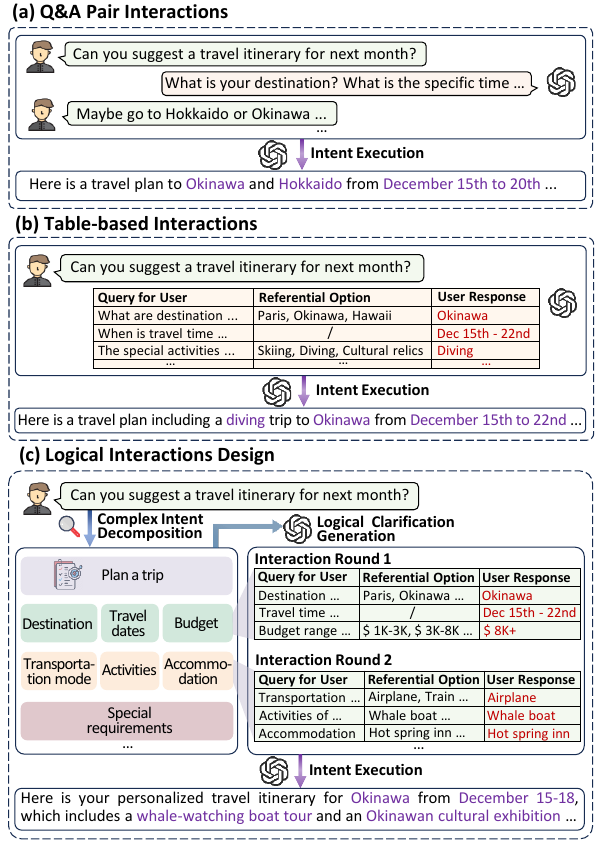}
\vspace{-20pt}
\caption{Comparison of user intent  clarification: (a) Q\&A Pair Interactions, (b) Table-based Interactions, and (c) Logical Interactions Design.}
\Description{}
\label{fig:compare}
  \vspace{-16pt}
\end{figure}

Recent studies have explored user-LLM interaction mechanisms that enable LLMs to proactively clarify user intents.
\mi employs sequential Q\&A pairs to iteratively elicit missing information~\cite{DBLP:journals/corr/abs-2402-09205}, as illustrated in Figure~\ref{fig:compare}(a).
Similarly, \itiu introduces a table-based interaction mechanism that generates multiple clarification questions simultaneously within a single turn~\cite{DBLP:conf/cikm/LiaoL024}, as shown in Figure~\ref{fig:compare}(b).
These pioneering approaches improve efficiency and interactivity, however, they implicitly assume that clarification questions are parallel and independent, i.e., a \emph{simple intent scenario}.

% The empirical results further indicate the significance of this problem (as shown in Figure~\ref{fig:comparison})
% This disparity underscores the central challenge of complex intent understanding: without modeling of logical dependencies among clarification questions, LLMs may generate clarification questions that lack coherence with real-world logic, leading to confusing or impractical interactions. 
% Therefore, effectively complex intent understanding requires ensuring that these \textbf{logic dependencies} are well-represented and maintained throughout the process.

In practice, many tasks involve \emph{complex intent scenarios}, where clarification questions exhibit logical dependencies.
As shown in Figure~\ref{fig:compare}(b), the model should confirm the \textit{``destination''} before suggesting possible \textit{``activities''}.
Failing to model these dependencies may lead to incoherent or impractical suggestions, such as \textit{``diving in Okinawa in December''}, when the weather is often freezing.
These recommendations will increase the cost of acquiring effective information for users, a phenomenon commonly referred to as \textit{cognitive load}~\cite{DBLP:journals/cogsci/Sweller88,sweller1998cognitive}.
% unsuitable for diving.
% not only increases user cognitive load but may also result in 
The observed decline in empirical performance of previous methods (vividly in Figure~\ref{fig:comparison}), when transferred from simple to complex intent scenarios, further demonstrates the intricacy of the task.
% Empirical results further indicate this limitation (Figure~\ref{fig:comparison}).
% When moving from simple to complex intent scenarios, performance improvements of \mi and \itiu drop sharply, underscoring their inability to handle logical dependencies.
Therefore, we initially identify \textit{modeling the logical dependencies} among clarification questions as the core challenge in complex intent understanding.
% the central challenge lies in modeling \textit{logical dependencies} among clarification questions to support coherent intent understanding.

% To respond to these challenges, we ground our design in Cognitive Load Theory (CLT)~\cite{DBLP:journals/cogsci/Sweller88}, which distinguishes \emph{intrinsic} load (complexity inherent to the task) and \emph{extraneous} load (overhead from suboptimal presentation). 
% We operationalize these principles in \our as follows.
% (1) Managing intrinsic load: We introduce a retrieving dataset (\cid) that provides schemas for complex intent decomposition.
% This allows vague, high-dimensional user intents to be decomposed into smaller, well-structured elements, simplifying the task and reducing its inherent complexity.
% (2) Minimizing extraneous load: To minimize extraneous load during interactions, \our ensures that clarification questions follow a logical prerequisite dependency sequence. 
% By presenting these questions in a structured order, we prevent users from having to reason about undecided variables and avoid logically conflicting prompts, making the process more intuitive.
% Intuitively, \our treats a user's complex intent like \textit{``white light''} and the \cid-driven schema as a \textit{``prism''}, which refracts it into a small set of hierarchical elements, forming a \textit{``colored spectrum''}. 
% This hierarchy then guides the logical clarification process, ensuring a smooth and low-friction user experience as users work through these elements.

% To address this challenge, we draw inspiration from
To systematically address the challenge, we draw upon Cognitive Load Theory (CLT)~\cite{sweller1994cognitive}, which academically distinguishes cognitive load into two types: \emph{intrinsic load}, stemming from the inherent complexity of a task; and \emph{extraneous load}, resulting from suboptimal task presentation or interaction design.
This perspective motivates us to design an approach, referred to as \our, that reduces the inherent complexity of user intent (intrinsic load) while simultaneously minimizing non-essential reasoning effort (extraneous load).
\our treats a complex user intent as \textit{``white light''}, and the logic-driven schema as a \textit{``prism''} that refracts it into hierarchical elements, forming a \textit{``colored spectrum''}.
These hierarchical elements guide the sequence (parallel or sequential) for clarification questions, thereby achieving a low-friction and logically coherent clarification process.
As shown in Figure~\ref{fig:compare}(c), \our first determines the \textit{``destination''}, then suggests relevant \textit{``activities''} in the subsequent turn, thereby generating practical recommendations such as \textit{``whale watching boat tour''}.

% Specifically, in line with CLT's emphasis on leveraging prior knowledge and structured information to reduce cognitive load, we first construct a complex intent decomposition (\cid) dataset spanning 27 domains and 429 intents. 
% \cid serves as prior knowledge to guide the LLM in decomposing user's complex intent into structured elements. 
% Building on this, \our sequences clarification questions according to the hierarchy of elements. 
% Clarification questions at the same level are grouped and presented together within a single interactive table, enabling logically coherent multi-turn interactions. 
% Once all elements are clarified, \our generates the final task-oriented response.
% To further enhance LLMs' capacity for logical intent clarification, we design an intent-aware reward (\ir) that holistically evaluates interaction quality (see Section~3). 
% Using Monte Carlo Tree Search (MCTS), we simulate large-scale user-LLM interactions to produce high-quality training data, which is then used to fine-tune the LLM with RL algorithms on IR.
% Through this pipeline, \our achieves logically coherent clarification. 
% For example, it first determines the \textit{``destination''} in an initial round and only then suggests relevant \textit{``activities''} in the next, ultimately yielding practical recommendations such as a \textit{``whale watching boat tour''}.

% Specifically, we propose \textbf{\our}, a novel framework that leverages CLT principles for logically coherent intent clarification.
Specifically, 
% \our operationalizes CLT in two ways: (1) Managing intrinsic load. We 
we first construct a retrieving dataset (\cid) that provides schema for complex intent decomposition across 27 domains and 429 intents.
This enables vague, high-dimensional intents to be decomposed into smaller, well-structured elements, reducing task complexity.
% (2) Minimizing extraneous load. We 
Second, we ensure that clarification questions are structured in accordance with logical prerequisite dependencies.
By sequencing clarification questions hierarchically, it prevents users from reasoning about undecided elements.
% , thereby optimizing the presentation of the interaction.
Third, an intent-aware reward and Monte Carlo Sample are introduced to simulate large-scale user-LLM interactions.
Last but not least, to guarantee practical and user-centric responses, we fine-tune the LLM using a self-evolved intent tuning approach to optimize logical clarification.
% , generating high-quality training data.
Extensive experiments are conducted to evaluate \our across three key dimensions: clarification interaction, intent execution, and cognitive load. 
Most notably, \our achieves SOTA logical clarification, reducing logical conflicts in clarification questions to 11.5\%. 
In addition, compared with \mi, \our enhances user experience by increasing interaction rating by 14.4\% and lowering average task time by 34.8\%. 
% Overall, these results highlight the advanced capability of \our in understanding complex intents and delivering coherent clarification.

To summarize, our contributions in this work are as follows:
\begin{itemize}
\item To the best of our knowledge, we are among the first to identify the key challenge of complex intent understanding in LLMs and propose a non-trivial solution;
\item We introduce CLT into this task and proposed a novel framework, \our, which reduces user cognitive load through logically coherent intent clarification; and
\item Extensive experiments demonstrate that \our enhances performance among all metrics, delivering coherent clarification in understanding complex intents.
% in clarification interaction, intent execution, and cognitive load, achieving SOTA performance in logical intent clarification, thereby promoting a new paradigm in complex intent understanding in LLMs.
\end{itemize}

\section{Related Work}
\label{sec: Related Work}
This section briefs relevant efforts from two perspectives.

\subsection{Non-interactive User Intent Understanding}
User intent understanding refers to the process by which LLMs infer users' true goals, typically achieved via non-interactive or interactive approaches~\cite{DBLP:conf/acl/ChengLZKWDTH24,DBLP:conf/acl/QianHZDQCZZL0024,DBLP:conf/cikm/LiaoL024}. 
Non-interactive intent understanding focuses on predicting user intents without direct feedback or dialog~\cite{DBLP:journals/corr/abs-2311-04155}. 
Existing methods can be broadly divided into human preference alignment and prompt optimization.
Human preference alignment adapts LLMs via supervised fine-tuning (SFT) and reinforcement learning with human feedback (RLHF), aligning model outputs with human intents~\cite{ouyang2022training}. 
Prompt optimization, in contrast, refines user inputs to improve intent prediction without modifying the underlying model~\cite{DBLP:journals/corr/abs-2311-04155, DBLP:journals/corr/abs-2306-16004}. 
For example, \citet{DBLP:journals/corr/abs-2311-04155} introduce a contrastive learning approach for Bayesian Prompt Optimization (\bpo), whereas \citet{DBLP:journals/corr/abs-2306-16004} propose a rewriting-based framework to disambiguate user queries.
Although effective, these methods primarily emphasize improving prediction accuracy rather than deepening or personalizing the understanding of user goals.

\subsection{Interactive User Intent Understanding}

Interactive intent understanding seeks to capture personalized user intents through effective user-model interaction~\cite{DBLP:journals/corr/abs-2402-09205,DBLP:conf/cikm/LiaoL024,DBLP:conf/icml/WuGP0LDC0L025}. 
For example, \citet{DBLP:journals/corr/abs-2402-09205} introduce \mi, an interactive model that proactively elicits missing user information via  Q\&A pairs exchanges.
\itiu~\cite{DBLP:conf/cikm/LiaoL024} propose a table-based interaction mechanism that generates multiple clarification questions in parallel to improve efficiency.
\collab~\cite{DBLP:conf/icml/WuGP0LDC0L025} incorporates a multiturn-aware reward function that emphasizes user engagement, optimizing LLMs toward user-centric clarification.
Despite these advances, existing methods overlook a crucial aspect—the logical sequencing of clarification questions.
Without modeling dependencies among questions, they struggle to handle the complex intent scenarios commonly observed in real-world contexts.
% In contrast, our approach explicitly models such logical dependencies, enhancing LLM adaptability to complex intents while reducing user cognitive load, leading to more intuitive and efficient interactions.

\begin{figure*}[!t]

    \centering
    \includegraphics[width=\linewidth]{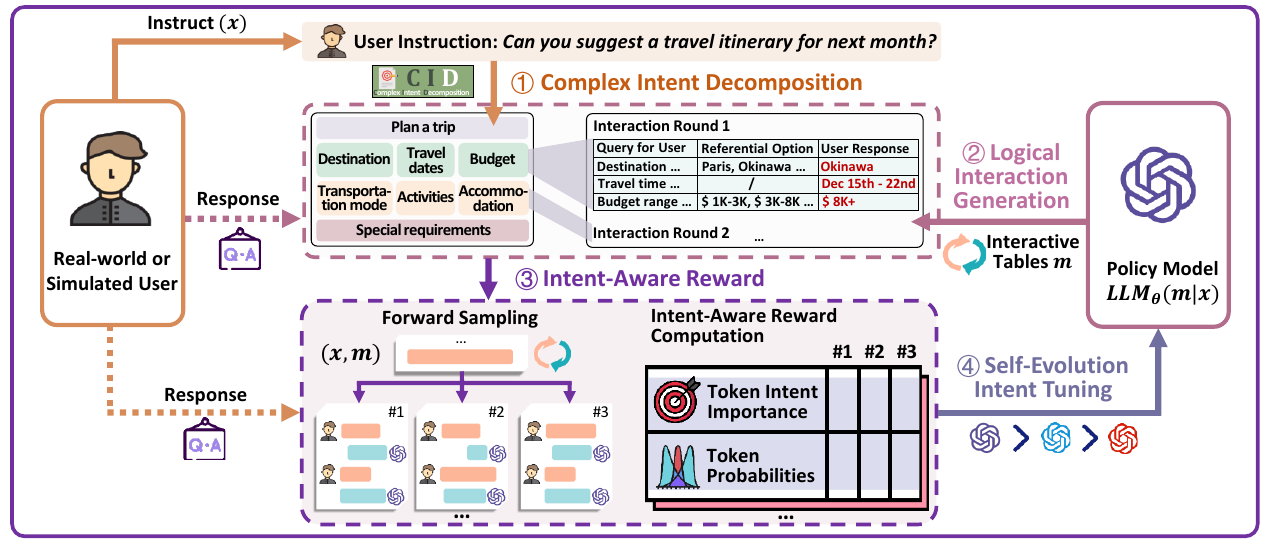}
    \vspace{-18pt}
    \caption{The \our Framework: Given a user instruction, the policy model first \circlednum{1} hierarchically decomposes complex intents with retrieving \cid dataset. 
It then organizes clarification questions and enables \circlednum{2} logical interaction through interactive table. 
\circlednum{3} Intent-aware rewards (IRs) are estimated via Monte Carlo sampling. 
Finally, \circlednum{4} self-evolved intent tuning iteratively enhances the training data quality and improves LLM's capacity for complex intent understanding.}
\Description{}
    \label{fig:framework}
    \vspace{-5pt}
\end{figure*}

\section{Problem Formulation}
\label{sec: Problem Formulation}
This section provides several formal definitions of key concepts that are essential to our work.

\myparagraph{Complex Intent}
Building upon previous studies~\cite{DBLP:journals/ijon/ZhouYWZS24,DBLP:conf/interspeech/BohusR03,tate1977generating,cohen1979elements}, we define a complex intent $z$ as one that (1) can be decomposed into multiple elements and 
% ~\cite{DBLP:conf/caise/BougueliaBBBZK22,DBLP:journals/ijon/ZhouYWZS24,DBLP:conf/interspeech/BohusR03,tate1977generating,cohen1979elements}
(2) exhibits prerequisite dependencies among these elements.  
For example, the intent \textit{``plan a trip''} includes elements such as \textit{``destination''}, \textit{``travel dates''}, and \textit{``activities''}.
Determining the \textit{``destination''} logically precedes choosing \textit{``activities''}, allowing the model to generate activities suggestions that are relevant to the selected location.

% , where determining the \textit{``destination''} should logically precede choosing specific \textit{``activities''}, enabling the model to suggest contextually relevant local activities.

\myparagraph{Multi-turn Clarification}
Given an initial instruction $x$, the model performs $K$ clarification turns before generating the final output $y$.  
At the $j$-th turn, conditioned on the $x$ and prior clarification history $t_{1:j-1}$, the model generates a clarification $m_j$ (e.g., an interactive table), to which the user provides a response $u_j$. 
The $j$-th clarification is denoted as $t_j = \{m_j, u_j\}$, and the overall interaction process is represented as $\{x, t_{1:K}, y\}$, where $t_{1:K} = \{t_1, \dots, t_K\}$ indicates the complete clarification trajectory.

\myparagraph{Objective}
The objective is to generate a clarification trajectory $\{t_{1:K}\}$ that is logical and coherent, ultimately leading to a high-quality final output $y$. 
The quality of the clarification trajectory and final output is evaluated by an intent-aware reward function, $R^*(t_{1:K}, y \mid x)$, which quantifies how effectively the clarification process enhances alignment between the output $y$ and the user intent.
Formally, the optimization objective is expressed as:
\[
\max_{y} \; R^*(t_{1:K}, y \mid x),
\]
which represents the goal of maximizing the alignment between the model's output and the user's intent.

\section{Proposed Method}

% explains the motivation behind the \our and then 
This section first provides an overview of \our framework with the detailed design of each module.

% \myparagraph{Motivations}
% Existing methods for interactive intent understanding, such as \mi~\cite{DBLP:journals/corr/abs-2402-09205} and \itiu~\cite{DBLP:conf/cikm/LiaoL024}, primarily focus on enabling LLMs to proactively ask clarification questions instead of directly executing ambiguous intents. 
% While these methods enhance proactive clarification, they neglect the logical dependencies inherent the user-LLM clarification process.
% This oversight leads to dependency conflicts across multi-turn conversations, $t_{1:K}$, hindering the efficient achievement of the objective $R^*(t_{1:K},y \mid x)$.
% For example, as illustrated in Figure~\ref{fig:compare}(b), clarification questions about \textit{``destination''} and \textit{``activities''} are asked in the same conversation turn.
% This prevents the model from recommending destination-specific activities, increases the user’s cognitive load, and results in impractical planning suggestions.
% To be effective, clarification should explicitly model and respect these logical dependencies, ensuring coherence and alignment with real-world reasoning. 
% Specifically, the model should account for the impact of these logical dependencies on the entire clarification trajectory $t_{1:K}$, thereby maximizing the intent-aware reward $R^*(t_{1:K},y \mid x)$.

\subsection{Framework}
In address to the identified challenges, we design four tailored modules guided by CLT.
% \ljz{Add some technical details into each module.}
The overall framework of \our is illustrated in Figure~\ref{fig:framework}.
Given a user instruction, the model first \circlednum{1}~decomposes complex intent into hierarchical elements by retrieving or few-shot constructing schema from \cid, ensuring prerequisite dependencies are preserved (ref. Section~\ref{sec: Complex Intent Decomposition}). 
Next, it \circlednum{2}~organizes and presents clarification questions hierarchically: independent questions are grouped into an interactive table within a single turn, while dependent questions are asked sequentially to maintain logical consistency (ref. Section~\ref{sec: Logical Interaction Generation}).
Subsequently, we simulate user-LLM clarification interactions via Monte Carlo Sampling, using an \circlednum{3}~intent-aware reward that integrates token-level intent importance and generation confidence to evaluate and filter clarification trajectories (ref. Section~\ref{sec: Intent-Aware Rewards}).
Finally, \circlednum{4}~the self-evolved intent tuning module iteratively refines the model through fine-tuning (SFT/DPO) on high-reward data, where each iteration generates higher-quality data for subsequent optimization rounds (ref. Section~\ref{sec: Self-Evolved Intent Tuning}).

\subsection{Complex Intent Decomposition}
\label{sec: Complex Intent Decomposition}
Complex intent decomposition aims to decompose a user's intent into smaller, well-structured elements, thereby reducing task complexity.
To support this process, we construct the \underline{c}omplex \underline{i}ntent \underline{d}ecomposition dataset (i.e., \cid), which provides annotated element dependencies for 429 common intents across 27 domains.

% This module incorporates real-world clarification logic to resolve logical dependency conflicts between clarification questions and aligns with cognitive load theory. 
% By leveraging structured information and prior knowledge, it minimizes cognitive load during user interactions, thus improving the efficiency of the clarification process.

\subsubsection{Construction of \cid Retrieval Dataset}
We construct the \cid retrieval dataset following two key principles:
(1) Multi-source integration with consistent naming to mitigate data heterogeneity, and
(2) Explicit prerequisite annotation to capture logical dependencies among elements.
The construction process comprises two stages: multi-source integration and dependency identification.
Both stages employ LLM-assisted annotation followed by human review, as detailed in Appendix~\ref{app:Construction Details of CID}.

\myparagraph{Multiple Sources Integration}
We leverage existing datasets from related tasks as data sources (see Appendix~\ref{app:Data Sources} for  statistics).
These datasets contain annotations for key components such as domain, intent, and elements.
Given their heterogeneity, we apply LLM-based semantic clustering to standardize synonymous expressions across datasets, thereby ensuring consistency.
Each cluster is assigned a canonical label, typically corresponding to the most frequent or standardized form.
For example, expressions such as \textit{``Plan a trip''} and \textit{``Travel planning''} are unified under the canonical intent \textit{``Plan a trip''}, with their associated elements merged accordingly.

\myparagraph{Dependency Identification}
We leverage LLM to assist human annotators in labeling logical dependencies among elements.
First, we instruct \textsf{GPT-4o} to predict prerequisite elements for each target element. 
Human annotators then validate these predictions by checking counterexamples (e.g., verifying whether \textit{``activities''} can be clarified without first specifying the \textit{``destination''}). 
Based on the identified prerequisites, elements are organized into hierarchical layers under two constraints:
(1) elements within the same layer are independent, and (2) dependencies flow strictly from earlier to later layers.
This hierarchical structure enables a clarification process in where independent clarification questions are posed in parallel within a single turn, while dependent questions are deferred to subsequent turns.
Further details of the annotation process are provided in Appendix~\ref{app:Annotation Process}.

Through these two stages, we construct the \cid retrieval dataset.
A distinctive feature of \cid is its explicit encoding of prerequisite relationships among intent elements, providing a reference schema for complex intent decomposition.

\subsubsection{Intent Decomposition Process}
Given a initial user instruction $x$, the goal of intent decomposition is to derive its hierarchical elements $L(x)=\{\ell_1,\dots,\ell_H\}$.
We adopt a \emph{retrieve-or-construct} paradigm based on the \cid dataset: if the user intent matches an existing intent in \cid, we reuse its annotated hierarchical elements; otherwise, we construct new hierarchical elements via few-shot generalization from the most similar intents in \cid.

\myparagraph{Domain and Intent Recognition}  
We first predict  the most probable domain-intent pair $(d^*,z^*)$ for the initial user instruction $x$.
Formally, the LLM estimates:
\begin{equation}
(d^*,z^*)=\arg\max_{(d,z)}\,LLM_\theta(d,z \mid x, D, Z),
\end{equation}
where $LLM_\theta$ denotes the probability distribution provided by the LLM, and $D$ and $Z$ are predefined domain and intent sets in \cid to ensure naming consistency. 
If $(d^*,z^*)$ exists in \cid, we directly reuse its annotated hierarchical elements; otherwise, we perform few-shot construction to generate them.

\myparagraph{Few-shot Construction}  
When no exact match is found, we retrieve the $k$ most semantically similar intents from \cid as exemplars\footnote{We apply the Contriever~\cite{DBLP:journals/tmlr/IzacardCHRBJG22} method to compute semantic similarity.}. 
% Exemplar $E$ is selected using embedding similarity.
% \begin{equation}
% E=\operatorname{argtop}_k^{i=1,\ldots,|Z|}\left[\mathbb{E}(z_i)^\top\cdot \mathbb{E}(z^*)\right],
% \end{equation}
% where $\mathbb{E}(\cdot)$ denotes the embedding function, and the dot product measures similarity. 
% These retrieved exemplars and their annotated hierarchical elements are used as few-shot examples $E$.
The retrieved exemplars, along with their annotated hierarchical elements, are provided as few-shot examples $E$.
The model then generates new hierarchical elements for $x$ conditioned on these examples:
\begin{equation}
L(x)=LLM_\theta(L\mid E,x,z^*).
\end{equation}
This process enables the LLM to transfer decomposition logic from existing intents in \cid to unseen ones, while maintaining logical consistency among dependencies.

\subsection{Logical  Clarification Generation}
\label{sec: Logical Interaction Generation}
This module organizes the sequence of clarification questions according to the element hierarchy $L(x)=\{\ell_1,\dots,\ell_H\}$, to facilitate logically coherent interactions with the user.
For each layer of elements $\ell_k$, the LLM generates an interactive table containing corresponding clarification questions and recommended options:
\begin{equation}
m_k= LLM_{\theta}(x, t_{1:k-1},\ell_k),
\end{equation}
where $t_{1:k-1}$ denotes the clarification history of the previous turns.

Clarification questions within the same layer $\ell_k$ are independent and thus presented together in a single table to enable parallel interaction.
User responses are denoted as $u_k$ and combined with the clarification history, i.e., $t_{1:k} = \{m_k,u_k\} \cup t_{1:k-1}$.
Clarifications for the next layer $\ell_{k+1}$ are then generated based on $t_{1:k}$, ensuring that the clarification trajectory adheres to the logical dependencies defined by the element hierarchy.
% This design allows users to resolve independent elements simultaneously while keeping dependent elements strictly sequential, thereby lowering cognitive load and maintaining real-world logic in multi-turn interactions.

\subsection{Intent-Aware Rewards}
\label{sec: Intent-Aware Rewards}

% In Figure~\ref{fig:framework}, our main insight is that effective complex intent understanding relies on logical intent clarification strategies. 
% Given the user's initial instruction, the model should consider real-world logical dependencies to organize the order of clarification questions. 
% To capture this, we design an intent-aware reward module to estimate the quality of clarification trajectory, and through an Monte Carlo sampling to obtain large-scale high-quality training data for LLM fine-tuning. 
To enhance the logical clarification capability of LLM, we employ Monte Carlo Sampling to generate large-scale high-quality training data for fine-tuning.
During this process, we design an intent-aware reward to evaluate the quality of each clarification trajectory.
The intent-aware reward (IR) for the clarification $m_j$ generated at the $j$-th turn is:
\begin{equation}
\begin{aligned}
&\text{IR}(m_j\mid x, t_j^h)\\
% &= ~ \mathbb{E}_{y\sim P(t_{j:K},y|t_{1:j-1}\cup\{m_j\},x)}R^*(t_{1:j-1}\cup\{m_j\}\cup t_{j:K},y\mid x)\\
&= ~ \mathbb{E}_{y\sim P(t_{j}^{f},y|x,t_j^h)}R^{*}(t_j^h\cup t_{j}^{f},y\mid x),
\end{aligned}
\label{eq:IR}
\end{equation}
where $t_j^{h}=t_{1:j-1} \cup \{m_j\}$ denotes the conversation history up to the $j$-th turn, $t_{j}^{f}={u_j} \cup t_{j+1:K}$ represents the forward  conversation after the $j$-th turn. 
The distribution $P(t_{j}^{f},y|x,t_j^h)$ models potential future conversations conditioned on the prior history.

However, computing Equation~\ref{eq:IR} remains challenging, as it requires: (1) a token-level reward function, $R^{*}(t,y|x)$ to evaluate each clarification trajectory $t$, and (2) a sampling strategy to obtain the forward conversation distribution $P(t_{j}^{f},y|x,t_j^h)$.
We elaborate on these two components in Section~\ref{sec: Intent-aware Reward Function} and~\ref{sec: Forward Sampling}.

\subsubsection{Token-level Reward Function}
\label{sec: Intent-aware Reward Function}
% To effectively estimate the quality of a clarification trajectory, 
We define the token-level reward $R^{*}(t,y \mid x)$ to jointly capture the importance of each token in expressing user intent and the model's confidence in generating it.
Given a clarification trajectory $t$ and a simulated final output $y$ obtained through forward sampling, we compute:
\begin{equation}
\begin{aligned}
&R^*(t,y\mid x)\simeq R_{\mathrm{imp}}(x,t,y)^\top \cdot R_{\mathrm{con}}(x,t,y), \\
&R_{\mathrm{imp}}(x,t,y)^\top, R_{\mathrm{con}}(x,t,y)\in\mathbb{R}^{N\times1},
\end{aligned}
\end{equation}
where $R_{\mathrm{imp}}(x,t,y)$ computes the intent importance score vector of $y$, and $R_{\mathrm{con}}(x,t,y)$ computes the generation confidence score vector for the $N$ tokens in $y$. 
Each vector dimension corresponds to a token in $y$, indicating how essential it is for expressing the user's intent and how confidently the model generates it.

For instance, in Figure~\ref{fig:compare}(c), the token \textit{``Okinawa''} is critical for expressing the  \textit{``destination''} intent and thus receives a high intent importance score.
If the model also generates this token with high confidence, the clarification trajectory $t$ effectively clarifies the \textit{``destination''} intent. 
Conversely, low generation confidence for such a key token indicates that the clarification process remains incomplete or inconsistent.

\myparagraph{Intent Importance Score} 
$R_{\mathrm{imp}}(x,t,y)$ measures the contribution of each token to the overall intent expressed in $y$.
We estimate intent importance by assessing the sensitivity of $y$ to semantic contradictions.
Specifically, we define a loss $\mathcal{L}$ that quantifies the likelihood of \textit{``contradiction''} between $y$  and itself using a natural language inference (NLI) model. 
While the model typically predicts \textit{``entailment''} with high probability, we invert this label and compute the gradient of the \textit{``contradiction''} loss with respect to each token embedding $\mathbf{e}_i$. 
The intent importance score is then computed as:
% \begin{equation}
% \begin{aligned}
% \mathbf{R}_{\mathrm{imp}}(x,t,y)&=\begin{bmatrix}\|\mathbf{e}_1\odot\nabla_{\mathbf{e}_1}L\|_2\\\|\mathbf{e}_2\odot\nabla_{\mathbf{e}_2}L\|_2\\\vdots\\\|\mathbf{e}_N\odot\nabla_{\mathbf{e}_N}L\|_2\end{bmatrix}\\
% &=\left[\left.\|\mathbf{e}_i\odot\nabla_{\mathbf{e}_i}L\|_2\right.\right]_{i=1}^N\in\mathbb{R}^{N\times1},
% \end{aligned}
% \end{equation}
\begin{equation}
R_{\mathrm{imp}}(x,t,y)=\left[\left.\|\mathbf{e}_i\odot\nabla_{\mathbf{e}_i}\mathcal{L}\|_2\right.\right]_{i=1}^N\in\mathbb{R}^{N\times1},
\end{equation}
where $\odot$ denotes element-wise multiplication. 
The $i$-th dimension of $R_{\mathrm{imp}}(x,t,y)$ reflects the sensitivity of token $y_i$ to perturbations in the overall intent representation.
A higher value indicates that the token is more critical for expressing the intent~\cite{DBLP:conf/nips/AdebayoGMGHK18}.

\myparagraph{Generation Confidence Score}
$R_{\mathrm{con}}(x,t,y)$ measures the LLM's confidence when generating final output $y$. 
For each token $y_i$, the LLM produces a conditional probability distribution given the preceding context $(x,t,y_{<i})$:
% We therefore define:
\begin{equation}
R_{\mathrm{con}}(x,t,y)=
\left[P(y_i \mid x,t,y_{<i})\right]_{i=1}^N \in\mathbb{R}^{N\times1},
\end{equation}
where $i$-th dimension of $R_{\mathrm{con}}(x,t,y)$ corresponds to the LLM's predicted probability for token $y_i$.
A higher probability indicates greater confidence, suggesting that the clarification trajectory $t$ provided sufficient information to generate $y_i$ reliably.

By integrating $R_{\mathrm{imp}}$ and $R_{\mathrm{con}}$, the token-level reward function highlights tokens that are both semantically important and confidently generated. 
This design ensures that the reward accurately reflects whether the clarification trajectory resolves key intent elements with logical consistency.

\subsubsection{Forward Sampling}
\label{sec: Forward Sampling}
To compute Equation~\ref{eq:IR}, we require samples from $P(t_{j}^{f},y \mid x,t_j^h)$—the distribution of forward conversation conditioned on the conversation history. 
A simple approach is to apply Monte Carlo Sampling, where the conversation is expanded turn-by-turn until completion.
Unlike standard Monte Carlo Sampling, we do not rely on an external reward model; instead, we use \ir to estimate the intent-aware reward score at each turn. 
This approach reduces computational overhead while providing a more accurate measure of intent awareness in the final output $y$.

\begin{table*}[!t]
\caption{The results of clarification interaction evaluation across various metrics on the \tin test set (see Table~\ref{tab:clarification interaction in3} and \ref{tab:clarification interaction abp} for the results of \inth and \abp). 
The comparisons include the \textsf{Interact} series, \itiu, \collab, and \our (with \textsf{SFT} and \textsf{DPO} variants), each utilizing two baseline model configurations.
Arrows represent the higher ($\uparrow$) or the lower ($\downarrow$) the better.
The best results are highlighted in bold.
}
\vspace{-10pt}
\begin{center}
\footnotesize
\tabcolsep=0.002\linewidth
\begin{tabular}{L{0.05\linewidth}L{0.20\linewidth}C{0.001\linewidth}|C{0.065\linewidth}C{0.06\linewidth}C{0.07\linewidth}|C{0.075\linewidth}C{0.075\linewidth}C{0.001\linewidth}|C{0.065\linewidth}C{0.06\linewidth}C{0.07\linewidth}|C{0.075\linewidth}C{0.075\linewidth}}
\toprule
\rowcolor{CadetBlue!20} 
\multicolumn{2}{c}{\textbf{\textsf{Baseline}}}&~& \multicolumn{5}{c}{\textbf{\textsf{Mistral-7B-Instruct-v0.3}}}  &   ~& \multicolumn{5}{c}{\textbf{\textsf{LLaMA-3.1-8B-Instruct}}}\\

% \cmidrule(r){1-2} \cmidrule(r){4-10}  \cmidrule(){12-18} 

\rowcolor{CadetBlue!20} 
\textbf{Test Data}  &\textbf{Metrics} &  ~& \textbf{\mi} & \textbf{\itiu} & \textbf{\textsf{\textsc{Collab-LLM}}} &  \textbf{\textsf{Prism-SFT}}  & \textbf{\textsf{Prism-DPO}}  &~& \textbf{\textsf{LLaMA-Interact}} & \textbf{\itiu} & \textbf{\textsf{\textsc{Collab-LLM}}} &  \textbf{\textsf{Prism-SFT}}  & \textbf{\textsf{Prism-DPO}}  \\

\midrule
    ~  & \textbf{$^{\uparrow}$\textbf{Intents Cover Rate} (\%)} &  ~&   $63.15$      &     $59.45$       &    $67.19$ &    $66.97$&    $\textbf{70.81}$ &    ~&    $68.27$  &   $62.26$&     $68.61$  & $68.24$ &    $\textbf{71.98}$ \\

\rowcolor{gray!10}
~& \textbf{$^{\downarrow}$Average Interaction Turns} &~&      $4.21$         &   $\textbf{1.25}$&    $4.66$&    $1.39$&    $1.53$ & ~&        $4.53$ &   $\textbf{1.23}$&    $4.97$   &  $1.44$   & $1.57$        \\

\textbf{\tin-}& \hspace{0.5em}\textbf{Average Questions Per Turn} & ~ &    $1.13$      &    $2.57$   &   $1.78$  &   $2.29$&    $2.17$ &  ~&     $1.18$   &    $3.01$&    $1.61$  & $2.11$  & $2.08$  \\

% \cmidrule(r){2-2} \cmidrule(lr){4-10} \cmidrule(l){12-18}  
\rowcolor{gray!10}
\textbf{\texttt{Simple}} & \textbf{$^{\uparrow}$Options Presenting Rate (\%)}&    ~&  $\textbf{83.19}$     &      $60.94$      &  $55.27$  &  $74.57$   &   $71.06$ &  ~&      $\textbf{87.45}$    &    $78.37$&    $61.09$  &   $80.67$  &  $76.44$  \\

~& \textbf{$^{\uparrow}$Options Reasonable Rate (\%)} &  ~&    $\textbf{96.81}$      &        $96.48$    &    $88.07$ &    $96.57$ &  $94.32$  &    ~&   $97.04$ &    $\textbf{98.04}$&     $90.05$ &   $97.82$   & $95.66$       \\
\rowcolor{gray!10}
~ & \hspace{0.5em}\textbf{Average Options Per Question} & ~ &      $2.64$        &   $4.08$ & $1.27$   &    $4.13$  &    $3.91$  &   ~&   $2.67$     &  $4.16$  &   $1.31$  & $4.08$  &  $4.01$       \\

\midrule

~  & \textbf{$^{\uparrow}$\textbf{Intents Cover Rate} (\%)} &  ~&    $44.17$      &     $46.78$       &    $45.61$  & $55.93$ &    $\textbf{58.28}$ & ~&    $46.24$  &   $49.18$&    $57.08$    &   $57.53$   & $\textbf{60.09}$  \\

\rowcolor{gray!10}
~& \textbf{$^{\downarrow}$Logical Conflict Rate (\%)}   &        ~&     $39.50$      &     $53.00$       &    $34.50$  &   $\textbf{12.50}$  &    $\textbf{12.50}$ &  ~&      $39.00$   &  $51.50$  &  $30.00$    &    $13.00$ &    $\textbf{11.50}$  \\

~& \textbf{$^{\downarrow}$Average Interaction Turns} &~&      $5.64$       &  $1.89$    &   $6.36$  &    $\textbf{1.58}$&    $1.61$  & ~&        $5.87$ &   $1.78$   &    $6.14$ &  $1.66$   & $\textbf{1.64}$       \\

\rowcolor{gray!10}
\multirow{-2}{*}{\textbf{\tin-}}& \hspace{0.5em}\textbf{Average Questions Per Turn} & ~ &     $1.35$      &       $2.86$     &    $1.91$  &    $3.05$  &    $3.24$  &  ~&      $1.24$   &   $3.38$   &    $1.51$  &   $3.69$  & $3.73$    \\

\multirow{-2}{*}{\textbf{\texttt{Complex}}} & \textbf{$^{\uparrow}$Options Presenting Rate (\%)}&    ~&  $\textbf{74.21}$     &      $54.18$      &   $42.90$ &    $70.90$  &    $71.28$ &  ~&    $70.96$   &    $70.51$&    $48.71$  &   $\textbf{71.53}$    & $69.78$   \\
\rowcolor{gray!10}
~& \textbf{$^{\uparrow}$Options Reasonable Rate (\%)} &  ~&    $70.15$      &        $65.07$   &    $72.94$ &    $\textbf{89.17}$  &    $87.43$  &    ~&   $73.95$  &  $69.77$&    $70.69$  &   $90.44$   & $\textbf{92.19}$     \\

~& \hspace{0.5em}\textbf{Average Options Per Question} & ~ &      $2.37$      &       $3.21$     &   $1.16$  &    $3.81$  &    $3.42$  &   ~&    $2.24$     &    $3.43$  &    $1.10$  &  $3.97$  & $3.85$     \\
\bottomrule
\end{tabular}
\end{center}
\vspace{-5pt}
\label{tab:clarefication interaction}
\end{table*}

While real-world conversations can be collected from human participants, sampling multiple forward conversations during training is costly and inefficient. 
To enable scalable training, we introduce a user simulator $U$ that prompts the LLM to emulate user behavior.
The model is instructed to mimic the user's prior language style and inject typical user behaviors~\cite{DBLP:journals/corr/abs-2411-10109}, enabling efficient and realistic forward sampling. 
% The user simulator has an implicit goal, which it attempts to achieve during the conversation. 
% This design simulates real-world scenarios where users may have evolving needs, limited background knowledge, or require clarification, naturally unfolding into multi-turn conversations~\cite{DBLP:journals/corr/abs-2411-10109}.

\subsection{Self-Evolved Intent Tuning}
\label{sec: Self-Evolved Intent Tuning}
Leveraging the token-level reward function and forward sampling strategy, we can compute the \ir of any clarification trajectory without relying a separate reward model. 
This motivates us to perform intent tuning for LLMs in a self-evolved manner. 
The self-evolved intent tuning process unfolds over multiple rounds, during which data evolution and model evolution mutually reinforce each other.
In the first round, a powerful closed-source model (\textsf{GPT-4o}) serves as the policy model to generate large-scale intent clarification dialogue records, and select high-quality data based on the \ir to fine-tune a smaller open-source LLM (\textsf{LLaMA-3.1-8B-Intruct}). 
In the subsequent rounds, the the fine-tuned LLM serves as the new strategy policy model to generate progressively higher-quality training data for the next iteration.
We collect 47k user instructions as data sources, with detailed statistics provided in Table~\ref{tab:statistics}.
Specifically, self-evolved intent tuning process consists of the following stages.

\myparagraph{Clarification Trajectory Collection}
We employ Monte Carlo Sampling to generate intent clarification trajectories. 
Each trajectory is evaluated using its IR score and filtered for use in Supervised Fine-Tuning (SFT) and Direct Preference Optimization (DPO).
For SFT, top-ranked trajectories are retained to form the clarification dialogue training data.
For DPO, pairs of ``selected'' and ``rejected'' clarifications are constructed by ranking clarification trajectories according to their IR scores.

\myparagraph{Self-Evolved Intent Tuning}
We select \textsf{LLaMA-3.1-8B-Instruct}\footnote{~\url{https://huggingface.co/meta-llama/Llama-3.1-8B-Instruct}} and \textsf{Mistral-7B-Instruct-v0.3}\footnote{~\url{https://huggingface.co/mistralai/Mistral-7B-Instruct-v0.3}} as the  backbone models for \our.
The number of self-evolved rounds is determined based on empirical experience and prior research findings~\cite{DBLP:conf/nips/ZhangZHYD024,DBLP:journals/corr/abs-2501-04519}.
Three rounds of self-evolved intent tuning are performed, each generating higher-quality data than the previous one.
In each iteration, we use Monte Carlo Sampling to generate intent clarification trajectories, fine-tune on high-IR samples, and deploy the resulting model as the new policy for subsequent rounds.
By continuously optimizing for higher IR values, the model learns to generate responses that enhance the effectiveness and coherence of clarification interactions.
This self-evolved process allows \our to generate scalable, high-quality data for complex intent understanding without human annotation, thereby supporting robust generalization across diverse tasks.

% In summary, Prism can generate scalable complex intent understanding datasets without the need for manual annotation, enabling it to generalize across different tasks. In Section~\ref{sec: Related Work}, we compare Prism with related interactive intent understanding methods and highlight its contributions in lowering cognitive load and enhancing intent understanding.

\section{Experiments}
% The LLM's intent understanding capability can be evaluated directly through user-model clarification interaction, indirectly through downstream intent execution performance, and measured from the user's perspective in terms of cognitive load. 
% Clarification interaction focuses on intention understanding itself, intent execution emphasizes intention understanding’s ultimate goal, and cognitive load focuses on the actual user experience. 

% To comprehensively evaluate the effectiveness of \our, we divide our experiments into three aspects: 
% (1) \textbf{Clarification Interaction}, which evaluates the model's ability to understand user intent during the clarification interaction; 
% (2) \textbf{Intent Execution}, which measures the model's performance in executing the intent after clarification; and 
% (3) \textbf{Cognitive Load}, which gauges the user's experience in terms of mental effort during intent understanding.

The capability of an LLM to understanding user intent  can be evaluated from three complementary perspectives: clarification interaction, intent execution, and cognitive load. 
Clarification interaction assesses how effectively the model understands and refines user intent during the clarification process.
Intent execution reflects the model's ability to accurately and efficiently complete the intended task.
Cognitive load quantifies the user's mental effort throughout the interaction.
Accordingly, our experiments are organized around these three dimensions.

% (1) \textbf{Clarification Interaction}, which evaluates the model's ability to understand user intent during the clarification process; 
% (2) \textbf{Intent Execution}, which measures the model's performance in executing the intent after clarification; and 
% (3) \textbf{Cognitive Load}, which gauges the user's experience in terms of mental effort during intent understanding.

\begin{table*}[t]
    \centering
    \footnotesize
    \tabcolsep=0.006\linewidth
    \caption{The results of intent execution evaluation across various metrics in two baseline model configurations. 
    \textit{Rel. Improv.} indicates the relative improvements of \textsf{Prism-DPO} over \textsf{Baseline}.
    The arrows and corresponding values show the magnitude of increase or decrease compared to \textsf{Baseline}.
    The best results are highlighted in bold.
    % Arrows represent the higher ($\uparrow$) or the lower ($\downarrow$) the better. 
    }
    \label{tab:intent execution}
    \vspace{-8pt}
    \begin{tabular}{L{0.12\linewidth}|C{0.075\linewidth}C{0.075\linewidth}C{0.075\linewidth}C{0.075\linewidth}C{0.075\linewidth}|C{0.075\linewidth}C{0.075\linewidth}C{0.075\linewidth}C{0.075\linewidth}C{0.075\linewidth}}
        \toprule
        \rowcolor{CadetBlue!20}  \multicolumn{1}{c|}{\textbf{Test Data}} & \multicolumn{5}{c|}{\textbf{Simple Intent Scenario}}
        & \multicolumn{5}{c}{\textbf{Complex Intent Scenario}}\\
        \rowcolor{CadetBlue!20}  \multicolumn{1}{c|}{\textbf{Method}} & \textbf{$^{\uparrow}$BLEU}
        & \textbf{$^{\uparrow}$Faithful}
        & \textbf{$^{\downarrow}$US}
        & \textbf{$^{\downarrow}$GS}
        & \textbf{$^{\downarrow}$TI}
        & \textbf{$^{\uparrow}$BLEU}
        & \textbf{$^{\uparrow}$Faithful}
        & \textbf{$^{\downarrow}$US}
        & \textbf{$^{\downarrow}$GS}
        & \textbf{$^{\downarrow}$TI}\\
\hline
        \rowcolor{CadetBlue!20} \multicolumn{1}{c|}{\textbf{\textsf{Baseline} (\textsf{Mistral})}} 
        &  $29.8$
        &  $44.6$
        &  $2.67$
        &  $0.57$
        &  $4.34$
        &  $27.4$
        &  $39.7$
        &  $3.44$
        &  $0.46$
        &  $5.48$
        \\

        \rowcolor{gray!10}       \hspace{0.1em}  \textbf{\mi}
        &   $35.4$\up{5.6}
        &   $41.4$\down{3.2}
        &   $3.11$\down{0.44}
        &   $0.63$\down{0.05}
        &   $3.17$\up{1.17}
        &   $29.1$\up{1.7}
        &   $38.2$\down{1.5}
        &   $3.98$\down{0.54}
        &   $0.69$\down{0.23}
        &   $5.04$\up{0.46} \\

        \hspace{0.1em}    \textbf{\itiu}
        &   $37.2$\up{7.4}
        &   $50.9$\up{6.3}
        &   $2.51$\up{0.16}
        &   $0.71$\down{0.14}
        &   $3.26$\up{1.08}
        &   $30.8$\up{3.4}
        &   $43.1$\up{3.4}
        &   $3.28$\up{0.16}
        &   $0.51$\down{0.05}
        &   $4.97$\up{0.51} \\

        \rowcolor{gray!10}     \hspace{0.1em}    \textbf{\collab}
        &   $36.7$\up{6.9}
        &   $45.8$\up{1.2}
        &   $2.43$\up{0.24}
        &   $0.46$\up{0.11}
        &   $3.61$\up{0.73}
        &   $31.2$\up{3.8}
        &   $42.9$\up{3.2}
        &   $3.59$\down{0.15}
        &   $0.53$\down{0.07}
        &   $5.21$\up{0.27} \\

        \hspace{0.1em} \textbf{\textsf{Prism-SFT}}
        &  $38.6$\up{8.8}
        &  $53.4$\up{8.8}
        &  $\textbf{2.08}$\up{0.59}
        &  $0.41$\up{0.16}
        &  $\textbf{3.04}$\up{1.30}
        &  $\textbf{36.8}$\up{9.4}
        &  $46.6$\up{6.9}
        &  $2.58$\up{0.86}
        &  $0.40$\up{0.06}
        &  $4.37$\up{1.11}\\

        \rowcolor{gray!10}\hspace{0.1em} \textbf{\textsf{Prism-DPO}} 
        &  $\textbf{40.2}$\up{10.4}
        &  $\textbf{53.7}$\up{9.1}
        &  $2.27$\up{0.40}
        &  $\textbf{0.38}$\up{0.19}
        &  $3.11$\up{1.23}
        &  $35.9$\up{8.5}
        &  $\textbf{48.1}$\up{8.4}
        &  $\textbf{2.36}$\up{1.08}
        &  $\textbf{0.37}$\up{0.09}
        &  $\textbf{4.26}$\up{1.22}\\

        \hspace{0.1em} \textbf{\textsf{Rel. Improv.}}
        &  $34.90\%$
        &  $20.40\%$
        &  $14.98\%$
        &  $33.33\%$
        &  $28.34\%$
        &  $31.02\%$
        &  $21.16\%$
        &  $31.40\%$
        &  $19.57\%$
        &  $22.26\%$\\
\hline

        \rowcolor{CadetBlue!20}  \multicolumn{1}{c|}{\textbf{\textsf{Baseline} (\textsf{LLaMA})}} 
        &  $32.2$
        &  $45.4$
        &  $2.60$
        &  $0.53$
        &  $4.27$
        &  $28.5$
        &  $40.0$
        &  $3.38$
        &  $0.49$
        &  $5.45$
        \\

         \rowcolor{gray!10}   \hspace{0.1em}       \textbf{\mi}
        &   $36.1$\up{3.9}
        &   $43.6$\down{1.8}
        &   $2.88$\down{0.28}
        &   $0.61$\down{0.08}
        &   $3.22$\up{1.05}
        &   $29.7$\up{1.2}
        &   $39.6$\down{0.4}
        &   $3.95$\down{0.57}
        &   $0.65$\down{0.16}
        &   $5.14$\up{0.31} \\

             \hspace{0.1em}    \textbf{\itiu}
        &   $38.0$\up{5.8}
        &   $49.7$\up{4.3}
        &   $2.47$\up{0.13}
        &   $0.70$\down{0.17}
        &   $3.23$\up{1.04}
        &   $30.5$\up{2.0}
        &   $42.8$\up{2.8}
        &   $3.25$\up{0.13}
        &   $0.51$\down{0.02}
        &   $5.00$\up{0.45} \\

         \rowcolor{gray!10}     \hspace{0.1em}     \textbf{\collab}
        &   $35.9$\up{3.7}
        &   $45.0$\down{0.4}
        &   $2.40$\up{0.20}
        &   $0.48$\up{0.05}
        &   $3.67$\up{0.60}
        &   $30.8$\up{2.3}
        &   $39.9$\down{0.1}
        &   $3.50$\down{0.12}
        &   $0.51$\down{0.02}
        &   $5.19$\up{0.26} \\

        \hspace{0.1em} \textbf{\textsf{Prism-SFT}}
        &  $38.7$\up{6.5}
        &  $52.0$\up{6.6}
        &  $\textbf{2.18}$\up{0.42}
        &  $0.38$\up{0.15}
        &  $\textbf{3.10}$\up{1.17}
        &  $\textbf{36.3}$\up{7.8}
        &  $45.9$\up{5.9}
        &  $2.55$\up{0.83}
        &  $0.38$\up{0.11}
        &  $4.34$\up{1.11}\\

         \rowcolor{gray!10} \hspace{0.1em} \textbf{\textsf{Prism-DPO}} 
        &  $\textbf{39.6}$\up{7.4}
        &  $\textbf{53.1}$\up{7.7}
        &  $2.33$\up{0.34}
        &  $\textbf{0.35}$\up{0.18}
        &  $3.12$\up{1.15}
        &  $35.7$\up{7.3}
        &  $\textbf{47.6}$\up{7.6}
        &  $\textbf{2.28}$\up{1.10}
        &  $\textbf{0.36}$\up{0.13}
        &  $\textbf{4.22}$\up{1.23}\\

        \hspace{0.1em} \textbf{\textsf{Rel. Improv.}}
        &  $22.98\%$
        &  $16.96\%$
        &  $13.08\%$
        &  $33.96\%$
        &  $26.93\%$
        &  $25.61\%$
        &  $19.00\%$
        &  $32.54\%$
        &  $26.53\%$
        &  $22.57\%$\\
        \hline
    \end{tabular}
    \vspace{-5pt}
\end{table*}

\subsection{Evaluation on Clarification Interaction}
Clarification interaction does not involve task execution; instead, it focuses on assessing the model's ability to clarify user intents during the interaction process.

\subsubsection{Test Datasets and Settings}

To ensure a comprehensive evaluation, we select test instances from three intent understanding datasets. Each dataset contains annotations for user instructions, missing intent details, clarification processes, and final outputs.

\begin{itemize}
\item \textbf{\tin}~\cite{DBLP:conf/cikm/LiaoL024} consists of 200 test instances, each human-annotated with the assistance of \textsf{GPT-4}.
\item \textbf{\inth}~\cite{DBLP:journals/corr/abs-2402-09205} comprises 108 test instances, this dataset includes a greater number of clarification questions.
\item \textbf{\abp}~\cite{DBLP:conf/emnlp/Zhang0RNC24} contains 319 test instances, spanning diverse domains and covering a wider range of intents.
\end{itemize}

Following the definition of complex intent (ref. Section~\ref{sec: Problem Formulation}), we manually categorize test instances into simple and complex intent scenarios with the assistance of an LLM.
Comprehensive test data statistics are provided in Appendix~\ref{app:Test Data Statistics}.
For each instruction, users interact with the model in an open-ended setting, during which the model actively seeks clarification regarding the user's intent.
To ensure diversity in responses, we recruit 20 graduate students to participate in the interactions and provide responses.
Participants first list their intents outlines and then express them under the model's guidance. 
The entire clarification process is recorded and evaluated using specific metrics.

\subsubsection{Baselines and Competitors}
% We compare \our against three previous SOTA interactive intent understanding methods: \textsf{Mistral-Interact}\textsuperscript{\ref{footnote:IN3}}~\cite{DBLP:conf/acl/QianHZDQCZZL0024}, \textsf{ITIU}\textsuperscript{\ref{footnote:ITIU}}~\cite{DBLP:conf/cikm/LiaoL024}, and \collab\footnote{~\url{https://github.com/Wuyxin/collabllm}}~\cite{DBLP:conf/icml/WuGP0LDC0L025}.
% To ensure a fair comparison and control for differences in backbone models, we standardize all methods by adopting the same baseline models (i.e., \textsf{Mistral-7b-Instruct-v0.3} and \textsf{LLaMA-3.1-8B-Instruct}).

We compare \our against three previous SOTA interactive intent understanding methods: \textsf{Mistral-Interact}~\cite{DBLP:conf/acl/QianHZDQCZZL0024}, \textsf{ITIU}~\cite{DBLP:conf/cikm/LiaoL024}, and \collab~\cite{DBLP:conf/icml/WuGP0LDC0L025}.
To ensure a fair comparison and control for differences in backbone models, we standardize all methods by adopting the same baseline models (i.e., \textsf{Mistral-7b-Instruct-v0.3} and \textsf{LLaMA-3.1-8B-Instruct}).

\subsubsection{Implementation Details}
\our performs self-evolved intent tuning based on baseline models. 
We train two model variants: \ours and \ourd, each fine-tuned on a pre-generated multi-turn clarification dataset guided by the intent-aware reward (ref. Section~\ref{sec: Intent-Aware Rewards}).
The entire training process is conducted on eight 80GB A100 GPUs and completed in 14.5 hours, with the final checkpoint used for subsequent evaluation.

\subsubsection{Metrics}
We utilize the metrics developed by \citet{DBLP:journals/corr/abs-2402-09205} to transform subjective user intents from user-model clarification interactions into objective numerical values, simplifying data analysis and comparison.
These include Intents Cover Rate, Average Interaction Turns, Average Questions Per Turn, Options Presenting Rate, Options Reasonable Rate, and Average Options Per Question. 
Additionally, we introduce a Logical Conflict Rate metric to measure logical conflicts between clarification questions. The specific metric calculations are detailed in Appendix~\ref{app:Metric Details1}.

\subsubsection{Results}
Based on the clarification interaction evaluation results on \tin presented in Table \ref{tab:clarefication interaction} (see Appendix~\ref{app:Results of IN3 and ABP} for the corresponding evaluations on \inth and \abp),
our analysis of \our is summarized as follows.

\myparagraph{More Logical Clarification}
% In our analysis, \our demonstrates a significant improvement in clarification logic, outperforming all competitors. 
% This is evidenced by two key performance metrics:
% (1) Lower Logical Conflict Rate. \our achieves the lowest logical conflict rate of 11.50\%, demonstrating its superior ability to maintain logical consistency during the clarification process. 
% The reduction in logical conflicts translates to more accurate clarifications that align with real-world expectations, making the interaction process smoother and more intuitive for users. 
% (2) Higher Options Presenting Rate. The highest reasonable option ratio, 92.19\%, indicating that \our is better at providing contextually relevant clarifications, further improving the quality of interaction.
% Moreover, this improvement in clarification logic has not sacrificed interaction efficiency. 
% Despite better logical consistency, \our still maintains a competitive advantage in the number of interaction turns, with a value of 1.58. 
% This model achieves a balance between logical depth and interaction efficiency, which are critical factors in user satisfaction and task completion.
\our exhibits significant improvements in clarification logic, outperforming all competing methods.
This superiority is demonstrated by two key metrics:
(1) Lower Logical Conflict Rate. \our achieves the lowest logical conflict rate of 11.50\%, highlighting its superior ability to maintain logical consistency throughout the clarification process.
The reduction in logical conflicts results in more accurate and contextually coherent clarifications, thereby ensuring smoother and more intuitive user interactions.
(2) Higher Option Presentation Rate. \our attains the highest reasonable option ratio of 92.19\%, indicating stronger contextual relevance and higher-quality clarifications.
Importantly, these gains in logical consistency do not compromise interaction efficiency—\our maintains a competitive average of 1.58 interaction turns.
Overall, the model achieves a balanced trade-off between logical depth and interaction efficiency, both of which are essential for enhancing user satisfaction and task completion.

\myparagraph{Superiority in Complex Intent Scenarios}
% \our achieves a greater advantage in complex intent scenarios. 
% Specifically, with \textsf{Mistral} as the baseline, when transitioning from simple to complex intents, the Intent Cover Rate for \mi and \collab decreases by 30.1\% and 32.1\% respectively, while \textsf{Prism-DPO} only decreases by 17.7\%. 
% At the same time, in Options Presenting Rate and Average Options Per Question, the drop for \itiu is 11.1\% and 21.3\%, respectively, while \textsf{Prism-DPO} even shows a 3.1\% increase in Options Presenting Rate and only a 12.5\% decrease in Average Options Per Question. 
% This demonstrates that \our is more effective than its competitors in handling complex intent scenarios, thanks to \our's tailored complex intent decomposition and logical clarification strategies.
\our demonstrates clear advantages in complex intent scenarios.
Specifically, using \textsf{Mistral} as the baseline, when transitioning from simple to complex intents, the Intent Cover Rate of \mi and \collab decreases by 30.1\% and 32.1\%, respectively, while \textsf{Prism-DPO} shows only a 17.7\% decrease.
Similarly, for the Option Presentation Rate and Average Options Per Question, \itiu experiences declines of 11.1\% and 21.3\%, respectively, whereas \textsf{Prism-DPO} even records a 3.1\% increase in Option Presentation Rate and only a 12.5\% reduction in Average Options Per Question.
These results indicate that \our handles complex intent scenarios more effectively than its competitors, benefiting from its tailored complex intent decomposition and logical clarification strategies.

\myparagraph{Extensive Generalization}
As shown in Table~\ref{tab:clarification interaction in3} and \ref{tab:clarification interaction abp}, the results on \inth and \abp are consistent with results on \tin, with \our consistently outperforming all competing methods.
This consistency demonstrates the excellent generalization ability of \our across datasets.
Furthermore, regardless of whether \textsf{Mistral} or \textsf{LLaMA} is used as the baseline, \our maintains superior performance, showcasing its scalability across diverse backbone models.
These results highlight \our's robust generalization capabilities, confirming its effectiveness in real-world applications involving diverse tasks and complex user intents.

\begin{figure*}[!t]
\centering
  
\includegraphics[width=1.0\linewidth]{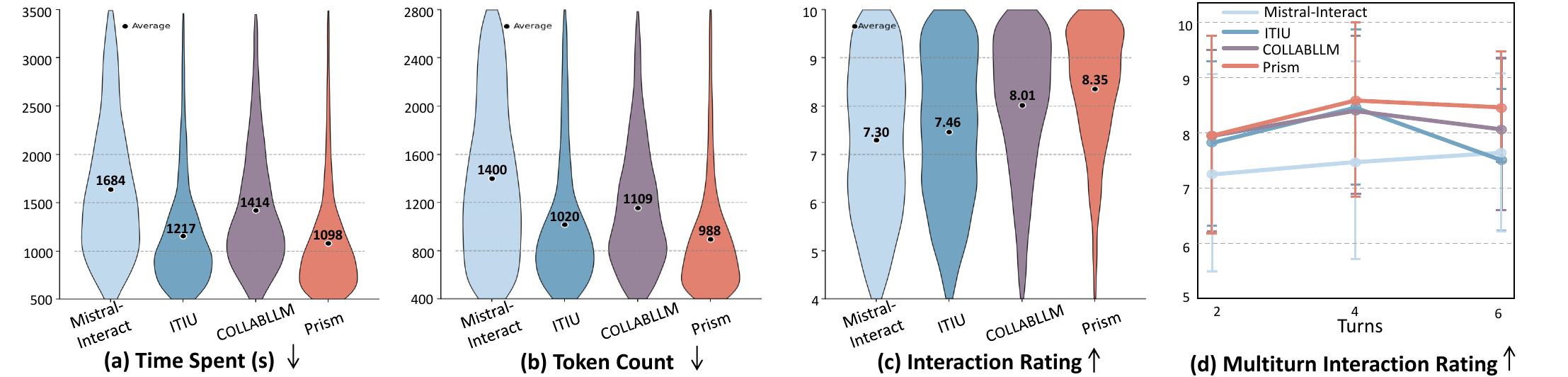}
\vspace{-20pt}
\caption{Our cognitive-load study includes 20 participants interacting with four anonymized models (\mi, \itiu, \collab, \our). Each participant is randomly assigned tasks from \tin, \inth, or \abp. 
We measure (a) user spent time, (b) conversation token count. Participants rate (c) overall interaction experience, with (d) additional assessments every two turns.}
\Description{}
\label{fig:load}
  \vspace{-13pt}
\end{figure*}

\begin{figure}[!t]
\centering

\includegraphics[width=1.0\linewidth]{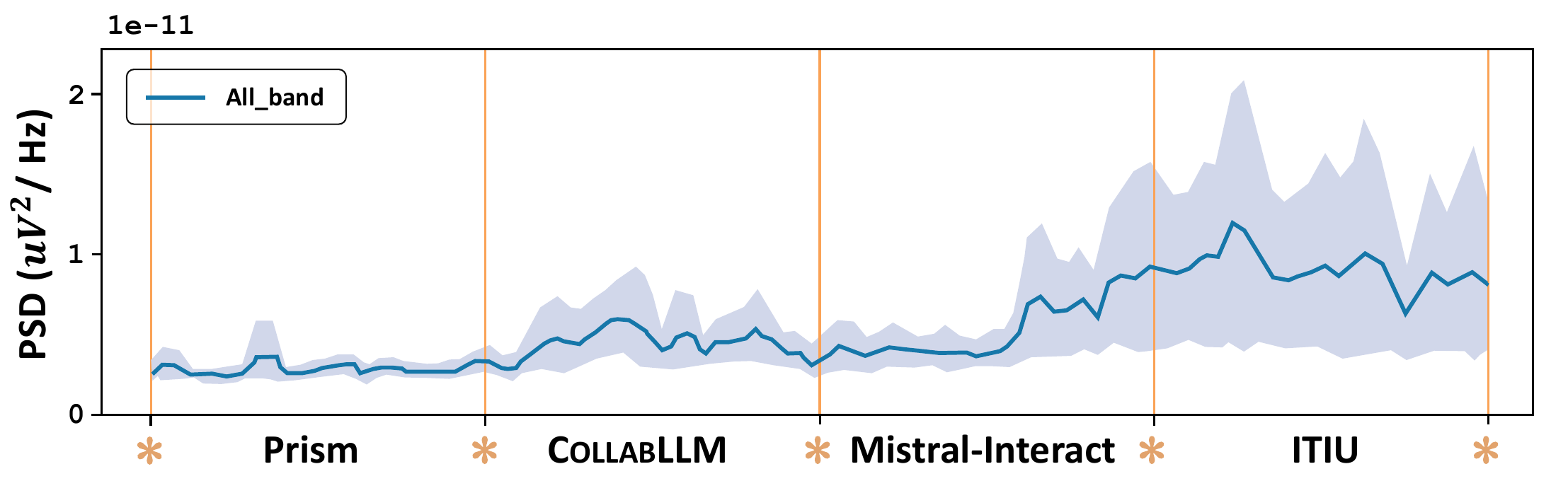}
  \vspace{-15pt}
\caption{Trend of absolute power spectral density (PSD) across the all band in the cognitive-load experiment. 
Each point on the blue solid line represents the mean PSD value of valid frequency points within the total frequency band at a given moment, and the shaded area indicates the standard deviation. 
Higher mean and variance of PSD indicate more active brain activity and higher cognitive load.}
\Description{}
\label{fig:brave}
  \vspace{-13pt}
\end{figure}

\subsection{Evaluation on Intent Execution}
\label{sec:Evaluation on Intent Execution}
To evaluate the effectiveness of user intent understanding from the perspective of intent execution, we integrate \our as upstream clarification interaction module within the \textsf{XAgent} framework~\cite{team2023xagent}, an autonomous agent system  for complex task solving. 
We conduct a proof-of-concept experiment by performance comparison.

\subsubsection{Test Datasets and Settings}
We randomly sample 100 tasks from \tin, \inth, and \abp that fall within the operational capabilities of \textsf{XAgent}.
For each task, participants first engage in clarification interactions with the target model, after which the interaction records are sent to \textsf{XAgent}.
\textsf{XAgent} then executes the user's intent based on these clarified records to generate the final output. 
During execution, \textsf{XAgent} divides the user's task into sub-tasks, which are completed through a series of tool invocations. 
We evaluate the intent execution process by measuring Unnecessary Sub-tasks (US), General Sub-tasks (GS), and Tool Invocation times (TI).
The quality of the final output is further assessed using the BLEU score and intent execution Faithfulness (Faithful).
Details of the metric computations are provided in Appendix~\ref{app:Metric Details2}.
For comparison, we include two baselines: \textsf{Mistral-7B-Instruct-v0.3} and \textsf{LLaMA-3.1-8B-Instruct}.
In baseline settings, \textsf{XAgent} directly executes the users' initial instructions without any prior clarification interactions.

% \subsubsection{Metrics}
% We use BLEU score, intent execution faithfulness (Faithful), unnecessary sub-tasks (US), general sub-tasks (GS), and tool invocation times (TI) as indicators to evaluate the effectiveness of this model in clarifying the user's task objectives and improving the efficiency of the intelligent agent's execution. The specific calculation methods of the indicators are presented in \todo{Appendix XX}.

\subsubsection{Results}
Table \ref{tab:intent execution} presents the results of the intent execution evaluation. 
Our findings regarding \our are summarized as follows: 
(1) Consistent performance across various intent scenarios. Under the \textsf{Mistral} baseline, when transitioning from simple to complex intent scenarios, the average relative improvement (\textit{Rel. Improv.}) of \textsf{Prism-DPO} across all metrics decreased slightly from 26.39\% to 25.08\%, a marginal reduction of 1.31\%. 
In contrast, \mi, \itiu, and \collab exhibited decreases of 13.64\%, 4.24\%, and 6.91\%, respectively. 
This stability highlights \our's strong robustness in handling complex user requirements.
(2) Enhanced logical planning ability. 
\textsf{Prism-DPO} achieves the fewest unnecessary and general sub-tasks, with averages of 2.27 and 0.38, outperforming all competitors.
This result indicates that effective upstream logical clarification substantially facilitates downstream agent planning.
(3) Improved tool execution efficiency. By reducing unnecessary and general sub-tasks, \our enhances the agent's efficiency in tool utilization, achieving an average of 3.04 tool invocations.
Overall, these results demonstrate that explicitly modeling user-specific intent and eliminating redundant actions enable \our to significantly improve agent performance in real-world applications.

\vspace{-3pt}
\subsection{Evaluation on Cognitive Load}
Cognitive load can be assessed via three types of measures: behavioral (user actions), subjective (users' self-reported ratings), and physiological (biometric data). 
To evaluate the model's effectiveness from a user-centric perspective, 
we conduct a user cognitive load study centered around 20 participants.

\subsubsection{Settings}
Each participant is assigned four random task drawn from the complex intent scenario in \tin, \inth, or \abp. 
% To simulate real-world scenarios where users have only a rough idea of the task, they are first asked to provide brief responses to topic-related questions.
Participants then sequentially interact with \our, \collab, \mi, and \itiu, without being informed of the model order in advance.
Every two turns, they provide an interaction rating based on their ongoing experience.
After completing the conversation, participants rate both the quality of model's final output and the overall interaction (subjective measures). 
All ratings are in a scale from 1 to 10. 
We also record the total interaction duration and total token count to assess behavioral efficiency (behavioral measures). 
Additionally, we use a multi-modal physiological signal acquisition system to record participants' Electroencephalogram (EEG) data, and measure their cognitive load through power spectral density (PSD) analysis (physiological measures).
Detailed experimental settings are provided in Appendix~\ref{app:Cognitive Load Experimental Details}.

\subsubsection{Results}
Figures~\ref{fig:load} and \ref{fig:brave} present the results of the cognitive load evaluation, with the analysis of \our as follows:
(1) More efficient user behavior. As shown in Figure~\ref{fig:load}(a), \our reduces task completion time by 34.8\% compared to \mi and by 22.3\% compared to \collab. 
In Figure~\ref{fig:load}(b), \our also lowers the total conversation token count to below 1,000, indicating that less information needed to be received and output by the user. 
This suggests that logical clarification plays a significant role in improving interaction efficiency.
(2) Higher quality user experience. In Figuree~\ref{fig:load}(c), \our consistently outperforms all competitors in user ratings, achieving an average interaction score of 8.35.
Specifically, 88.6\% of participants rate \our as ``good'' (score 8-9), and 58.1\% as ``very good'' (score 8-9), compared to 76.4\% and 34.2\% for \mi, respectively. 
In Figure~\ref{fig:load}(d), \itiu and \collab exhibit declining rating trends from turns 4–6, indicating reduced user satisfaction in longer interactions.
In contrast, \our’s ratings increase over time, demonstrating that its logically coherent interaction design effectively enhances the overall conversational experience.
(3) Lower user cognitive load. As shown in Figure~\ref{fig:brave}, the PSD gradually increases as participants interact sequentially with \our, \collab, \mi, and \itiu.
\our maintains the lowest PSD value, suggesting that users exert the least cognitive effort when interacting with our framework—consistent with its user-centered design objective.

\vspace{-8pt}
\section{Conclusions}

Understanding complex user intents is crucial for developing truly collaborative LLMs. 
While prior methods have improved efficiency and interactivity in simple intent scenarios, they struggle when clarification questions are logically dependent, forcing users to manage these relationships themselves and thereby increasing cognitive load.
The central insight of \our is that effective intent understanding requires explicitly modeling these dependencies among clarification questions to minimize the user's mental effort. 
Grounded in Cognitive Load Theory, \our decomposes ambiguous intents into smaller, well-structured elements and sequences clarification in a logically coherent manner, yielding smooth, low-friction interactions.
Extensive experiments and real-world evaluations demonstrate that \our achieves state-of-the-art logical clarification performance, enhancing user satisfaction and reducing task completion time.
\our advances LLMs toward being thoughtful, proactive collaborators that understand and guide users through complex, evolving goals.

\begin{acks}
This work was partially supported by 
National Natural Science Foundation of China (Nos. U25B2047, 62272469, 72301284), and
The Science and Technology Innovation Program of Hunan Province (No. 2023RC1007).
\end{acks}

\bibliographystyle{ACM-Reference-Format}
\balance
\bibliography{sample-base}

\appendix

\section{Interactive Method Comparison}
\label{app:Interactive method comparison}

\begin{table}[!h]
  \centering
  \footnotesize
  \tabcolsep=0.007\linewidth
  \vspace{-5pt}
  \caption{Compare \our with selected methods. (1) Versatile Interaction, evaluating its ability to support diverse  intent discovery beyond predefined behaviors; (2) Efficiency Optimization, evaluating whether efficiency optimization is performed; (3) User-Centric, determining whether user engagement or cognitive load, are explicitly considered; and (4) Logical Clarification, measuring whether it model the logical dependencies among clarification questions.}
  \label{tab:contribution}
  \vspace{-7pt}
  \begin{tabular}
    {L{0.26\linewidth}C{0.16\linewidth}C{0.19\linewidth}C{0.15\linewidth}C{0.17\linewidth}}
    \toprule
    \textbf{Method}  & \textbf{Versatile Interaction} & \textbf{Efficiency Optimization} & \textbf{User-Centric} & \textbf{Logical Clarification}\\
     % & Task-Agnostic & Broad Interaction Types & User-Centric Design & Causal and Objective-Aligned Reward \\
     \midrule
     \textbf{\bpo}~\cite{DBLP:conf/acl/ChengLZKWDTH24}
     & \large \textcolor{purple}{\ding{55}} 
     & \large \textcolor{purple}{\ding{55}} 
     & \large \textcolor{purple}{\ding{55}}
     &  \large \textcolor{purple}{\ding{55}}\\
     \textbf{\mi}~\cite{DBLP:journals/corr/abs-2402-09205}
     & \large \textcolor{teal}{\ding{52}} 
     & \large \textcolor{purple}{\ding{55}} 
     & \large \textcolor{purple}{\ding{55}}
     &  \large \textcolor{purple}{\ding{55}}\\
     \textbf{\itiu}~\cite{DBLP:conf/cikm/LiaoL024}
     &  \large \textcolor{teal}{\ding{52}}
     & \large \textcolor{teal}{\ding{52}} 
     & \large \textcolor{purple}{\ding{55}}
     & \large \textcolor{purple}{\ding{55}}\\
     \textbf{\collab}~\cite{DBLP:conf/icml/WuGP0LDC0L025}
     & \large \textcolor{teal}{\ding{52}}
     & \large \textcolor{teal}{\ding{52}}
     & \large \textcolor{teal}{\ding{52}} 
     & \large \textcolor{purple}{\ding{55}}\\
     \textbf{\our}
     &\large \textcolor{teal}{\ding{52}}
     &\large \textcolor{teal}{\ding{52}}
     &\large \textcolor{teal}{\ding{52}}
     &\large \textcolor{teal}{\ding{52}}\\
    \bottomrule
  \end{tabular}
  \vspace{-7pt}
\end{table}

In Table~\ref{tab:contribution}, we compare \our with related methods across four key dimensions. 
\our is a general, user-centric, and logical clarification framework.
% that explicitly models the logical dependencies of clarification questions to enable smoother interactions with users, thereby reducing their cognitive load.

\section{Construction Details of  \cid}
\label{app:Construction Details of CID}
\subsection{Data Sources}
\label{app:Data Sources}

% This integration broadens domain coverage and intent diversity while maintaining consistent annotation formats across heterogeneous sources.

% The combined dataset comprises 429 unique intents spanning 27 domains.
% Through normalization, deduplication, and semantic clustering, we ensure consistent representation of each intent, forming a unified and comprehensive foundation for complex intent decomposition in \cid.

\begin{table}[!h]
    \centering
\footnotesize
\vspace{-5pt}
    \caption{Statistics of \cid and its data sources. These datasets belong to Intent Understanding, Task-Oriented Dialogues, and Natural Language Understanding areas.
    For each dataset, we report the number of domains, intents, and elements, the size of the training set, as well as whether the annotations include logical dependencies among elements.}
    \vspace{-7pt}
    \label{tab:statistics}
    \tabcolsep=2pt
    \begin{tabular}{L{0.23\linewidth}C{0.07\linewidth}C{0.13\linewidth}C{0.09\linewidth}C{0.13\linewidth}C{0.12\linewidth}C{0.10\linewidth}}
    \toprule
    \textbf{Dataset} & \textbf{Area} & \textbf{\#Domain} & \textbf{\#Intent} & \textbf{\#Element} & \textbf{\#Size} & \textbf{Dependency}\\
    \midrule
    
       \textbf{\tin}~\cite{DBLP:conf/cikm/LiaoL024}  & ~ & 27 & 250 & 3,812 & 2,483 &\textcolor{purple}{\ding{55}}\\
      \textbf{\inth}~\cite{DBLP:journals/corr/abs-2402-09205}  & ~ & 10 & 250 & 3,615 & 1,261&\textcolor{purple}{\ding{55}} \\
    \textbf{\abp}~\cite{DBLP:conf/emnlp/Zhang0RNC24}  &   \multirow{-3}{*}{\textbf{IU}} & 10 & 328 & 1,800 & 1,000&\textcolor{purple}{\ding{55}}  \\
    \midrule
            
       \textbf{\texttt{SGD}}~\cite{rastogi2020towards} & ~  & 20 & 87 & 913 & 16,000&\textcolor{purple}{\ding{55}} \\
     \textbf{\texttt{ATIS}}~\cite{DBLP:conf/interspeech/MesnilHDB13} & ~  & 1& 26 & 127 & 4,978&\textcolor{purple}{\ding{55}} \\
     % \textbf{\texttt{SNIPS}}~\cite{DBLP:journals/corr/abs-1805-10190} &   & 12 & 7 & 72 & 13,784  \\
     \textbf{\texttt{MultiWOZ 2.1}}~\cite{DBLP:conf/lrec/EricGPSAGKGKH20}     &\multirow{-3}{*}{\textbf{TOD}} & 7 & 217 & 5,140 & 8,438&\textcolor{purple}{\ding{55}} \\
     \midrule

     \textbf{\texttt{MASSIVE}}~\cite{DBLP:conf/interspeech/MesnilHDB13} &  & 18 & 60 & 55 & -&\textcolor{purple}{\ding{55}} \\
            \textbf{\texttt{SNIPS NLU}}~\cite{coucke2018snips} & \multirow{-2}{*}{\textbf{NLU}}   & 7& 7 & 72 & 13,784 &\textcolor{purple}{\ding{55}}\\
         \midrule

     \textbf{\cid} & \textbf{IU}  & 27 & 429 & 3,714 & -&\textcolor{teal}{\ding{52}} \\

    \bottomrule
    \end{tabular}
\vspace{-7pt}
\end{table}

To construct the \cid retrieval dataset, we integrate data from multiple public benchmarks across three key research areas—Intent Understanding, Task-Oriented Dialogue, and Natural Language Understanding—as summarized in Table~\ref{tab:statistics}.

\subsection{Annotation Process}
\label{app:Annotation Process}
% For the multi-source integration stage, we adopted an LLM-assisted human annotation pipeline to ensure consistency and accuracy across heterogeneous datasets. The LLM first produced draft annotations by normalizing synonymous intent names and element sets. Three graduate annotators then refined these drafts, unified terminology, and corrected potential LLM biases. Each record was independently reviewed by a secondary annotator, and any disagreements were resolved through consensus. This hybrid approach combined LLM efficiency with human precision, achieving both scalability and reliability. The full process required about 84 working hours, covering automatic drafting, manual refinement, and cross-review.

% In the dependency identification stage, the LLM generated preliminary prerequisite relations among elements based on semantic and task logic. Human annotators validated these relations through counterexample reasoning—examining whether an element could be clarified independently of another—and removed cyclic or inconsistent dependencies. Final checks were performed by a senior reviewer to ensure domain-wide logical coherence. This LLM-human collaborative strategy yielded interpretable and accurate dependency structures, with the entire process taking approximately 144 working hours including drafting, verification, and quality assurance. 
% An illustrative annotation example is shown in Table~\ref{tab:cid-example}.

In the multi-source integration stage, we employed an LLM-assisted human annotation pipeline to ensure consistency and accuracy across heterogeneous datasets.
The LLM first produced draft annotations by normalizing synonymous intent names and element sets.
Three graduate annotators then refined these drafts, standardized terminology, and corrected potential LLM biases.
Each record was independently reviewed by a secondary annotator, and disagreements were resolved through consensus.
This hybrid approach combined LLM efficiency with human precision, achieving both scalability and reliability.
The full process required approximately 84 working hours. 
% encompassing automatic drafting, manual refinement, and cross-review.

In the dependency identification stage, the LLM generated preliminary prerequisite relations among elements based on semantic and task logic.
Human annotators validated these relations through counterexample reasoning—examining whether an element could be clarified independently of another—and removed cyclic or inconsistent dependencies.
A senior reviewer conducted final checks to ensure domain-wide logical coherence.
This LLM–human collaborative process yielded interpretable and accurate dependency structures, requiring roughly 144 working hours.
% for drafting, verification, and quality assurance.
An illustrative annotation example is provided in Table~\ref{tab:cid-example}.

\begin{table}[!h]
  \centering
  \footnotesize
  \setlength{\tabcolsep}{3pt}
  \vspace{-5pt}
  \caption{Example annotation record from the \cid dataset.
Each record contains the domain, intent, its elements, and the corresponding prerequisite relations among elements, from which hierarchical layers are induced.}
\vspace{-5pt}
  \label{tab:cid-example}
  \begin{tabular}{L{0.11\linewidth} L{0.17\linewidth} p{0.32\linewidth} p{0.29\linewidth}}
    \toprule
    \textbf{Domain} & \textbf{Intent ($z$)} & \textbf{Elements ($E$)} & \textbf{Prerequisites $\mathcal{P}(e_i)$} \\
    \midrule
    Travel 
    & Plan a trip 
    & \makecell[l]{%
      $e_1$: Destination\\
      $e_2$: Travel dates\\
      $e_3$: Budget\\
      $e_4$: Transportation mode\\
      $e_5$: Activities\\
      $e_6$: Accommodation\\
      $e_7$: Special requirements\\
    }
    & \makecell[l]{%
      $\mathcal{P}(e_1)=\varnothing$\\
      $\mathcal{P}(e_2)=\varnothing$\\
      $\mathcal{P}(e_3)=\varnothing$\\
      $\mathcal{P}(e_4)=\{e_1,e_2,e_3\}$\\
      $\mathcal{P}(e_5)=\{e_1,e_2,e_3\}$\\
      $\mathcal{P}(e_6)=\{e_1,e_2,e_3\}$\\
      $\mathcal{P}(e_7)=\{e_1,e_2,\dots,e_7\}$\\
    }\\
    \midrule
    \multicolumn{4}{l}{\textbf{Induced hierarchical layers} $L=\{\ell_1,\ell_2,\ell_3,\ell_4\}$} \\
    \multicolumn{4}{p{0.95\linewidth}}{%
      \makecell[l]{%
      $\ell_1$ (Intent): $z$\\
      $\ell_2$ (no prerequisites): $\{e_1,e_2,e_3\}$\\
      $\ell_3$ (all prereqs in $\ell_{\le 2}$): $\{e_4,e_5,e_6\}$\\
      $\ell_4$ (all prereqs in $\ell_{\le 3}$): $\{e_7\}$%
      }
    }\\
    \bottomrule
  \end{tabular}
  \vspace{-15pt}
\end{table}

\section{Clarification Interaction Experimental Details}
\label{app:Clarification Interaction Experimental Details}
\subsection{Training Details}
\label{app:Training Details}
We adopt \textsf{LLaMA-3.1-8B-Instruct} and \textsf{Mistral-7B-Instruct-v0.3} as the backbone models for \our and perform full-parameter and DPO fine-tuning for 5 epochs on 8×80GB A100 GPUs using the Hugging Face Transformers library~\cite{DBLP:conf/emnlp/WolfDSCDMCRLFDS20}.
Training employs the AdamW optimizer~\citep{DBLP:conf/iclr/LoshchilovH19} with $\beta_1=0.9$ and $\beta_2=0.999$, a learning rate of 1e-6, 10% warm-up steps, and linear decay scheduling.
We set the maximum input length to 4,096 tokens and use a batch size of 4.
The full training process completes in 14.5 hours, after which the final checkpoint is used for evaluation.

\subsection{Test Data Statistics}

\label{app:Test Data Statistics}
\begin{table}[!h]
\caption{Detailed statistics for the training and test splits of the \tin, \inth and \abp datasets.}
\vspace{-10pt}
\centering
\footnotesize
% \tabcolsep=0.02\linewidth
\tabcolsep=4.5pt
\begin{tabular}{lcccccc}
\toprule
\textbf{Dataset} & \multicolumn{2}{c}{\textbf{\tin}} & \multicolumn{2}{c}{\textbf{\inth}} & \multicolumn{2}{c}{\textbf{\abp}}\\
\midrule
\textbf{Split} & \textbf{Training} & \textbf{Test} & \textbf{Training} & \textbf{Test}& \textbf{Training} & \textbf{Test}\\
\cmidrule(){1-1} \cmidrule(l){2-3} \cmidrule(l){4-5}\cmidrule(l){6-7}
\begin{tabular}[c]{@{}l@{}}\textbf{Instruction}\\ \quad\quad- \textit{Complex}\\ \quad\quad- \textit{Simple} \end{tabular} 
& \begin{tabular}[c]{@{}c@{}}2,483\\ 2,143\\ 340 \end{tabular}
& \begin{tabular}[c]{@{}c@{}}200\\ 146\\ 54 \end{tabular}
& \begin{tabular}[c]{@{}c@{}}1,261\\ 1,012\\ 249 \end{tabular}
& \begin{tabular}[c]{@{}c@{}}108\\ 95\\ 13 \end{tabular}
& \begin{tabular}[c]{@{}c@{}}333\\ 207\\ 126 \end{tabular}
& \begin{tabular}[c]{@{}c@{}}319\\ 195\\ 124 \end{tabular}\\
\midrule
\begin{tabular}[c]{@{}l@{}}\textbf{\# Queries}\\ \quad\quad- \textit{Avg.}\\ \end{tabular}
& \begin{tabular}[c]{@{}c@{}}27,234\\ 12.71\\ \end{tabular}
& \begin{tabular}[c]{@{}c@{}}2,258\\ 15.47\\ \end{tabular}
& \begin{tabular}[c]{@{}c@{}}3,615\\ 3.57\\ \end{tabular}
& \begin{tabular}[c]{@{}c@{}}350\\ 3.68\\ \end{tabular}
& \begin{tabular}[c]{@{}c@{}}1,204\\ 3.62\\ \end{tabular}
& \begin{tabular}[c]{@{}c@{}}1,183\\ 3.71\\ \end{tabular}\\
\midrule
\begin{tabular}[c]{@{}l@{}}\textbf{\# Options}\\ \quad\quad- \textit{Avg.}\end{tabular}
& \begin{tabular}[c]{@{}c@{}}104,522\\ 3.84\end{tabular}
& \begin{tabular}[c]{@{}c@{}}8,499\\ 3.76\end{tabular}
& \begin{tabular}[c]{@{}c@{}}11,523\\ 3.19\end{tabular}
& \begin{tabular}[c]{@{}c@{}}1,042\\ 2.98\end{tabular}
& \begin{tabular}[c]{@{}c@{}}9,427\\ 2.61\end{tabular}
& \begin{tabular}[c]{@{}c@{}}1,166\\ 3.66\end{tabular}\\
\bottomrule
\end{tabular}
\label{tab:dataset}
\vspace{-15pt}
\end{table}

Table~\ref{tab:dataset} presents detailed statistics for the training and test splits of the \tin, \inth, and \abp datasets.
All test data are strictly excluded from every processing stage and used solely for evaluation.

\subsection{Metric Details}
\label{app:Metric Details1}
The metrics used for clarification interaction evaluation are as follows:
\begin{itemize}
\item \textbf{Intents Cover Rate}: It measures the percentage of underlying fact clarification questions that are covered by clarification questions generated by the model during the interaction.
\item \textbf{Logical Conflict Rate}: Indicates the proportion of clarification questions that violate prerequisite dependencies among intent elements, where lower values reflect more logically consistent and coherent clarifications.
\item \textbf{Average Interaction Turns}: It calculates the average number of interaction turns per instruction. 
\item \textbf{Average Questions Per Turn}: This metric determines the average number of clarification questions per turn of interaction.
\item \textbf{Options Presenting Rate}: This metric evaluates the percentage of the clarification questions accompanied by potential referential options.
\item \textbf{Options Reasonable Rate}: It records the percentage of referential options provided by the model that are considered reasonable.
\item \textbf{Average Options Per Query}: This represents the average number of referential options the model provides per query.
\end{itemize}

For all metrics, we apply direct statistical calculations or utilize \textsf{GPT-4} to assist in the measurements.
Additionally, we present only the macro-average results for all test metrics, as the micro-average results generally display similar trends across models.

\subsection{Results of \inth and \abp}
\label{app:Results of IN3 and ABP}
As shown in Table~\ref{tab:clarification interaction in3} and \ref{tab:clarification interaction abp}, the clarification interaction evaluation results on the IN3 and ABP datasets are consistent with those on TIN. Prism demonstrates the strongest logical interaction capability, adaptability to complex intent scenarios, and generalization ability.

\begin{table*}[!h]
\caption{The results of clarification interaction evaluation across various metrics on the \inth test set. 
The comparisons include the \textsf{Interact} series, \itiu, \collab, and \our (with \textsf{SFT} and \textsf{DPO} variants), each utilizing two baseline model configurations.
Arrows represent the higher ($\uparrow$) or the lower ($\downarrow$) the better.
The best results are highlighted in bold.
}
\vspace{-12pt}
\begin{center}
\footnotesize
\tabcolsep=0.002\linewidth
\begin{tabular}{L{0.05\linewidth}L{0.20\linewidth}C{0.001\linewidth}|C{0.065\linewidth}C{0.06\linewidth}C{0.07\linewidth}|C{0.075\linewidth}C{0.075\linewidth}C{0.001\linewidth}|C{0.065\linewidth}C{0.06\linewidth}C{0.07\linewidth}|C{0.075\linewidth}C{0.075\linewidth}}
\toprule
\rowcolor{CadetBlue!20} 
\multicolumn{2}{c}{\textbf{\textsf{Baseline}}}&~& \multicolumn{5}{c}{\textbf{\textsf{Mistral-7B-Instruct-v0.3}}}  &   ~& \multicolumn{5}{c}{\textbf{\textsf{LLaMA-3.1-8B-Instruct}}}\\

\rowcolor{CadetBlue!20} 
\textbf{Test Data}  &\textbf{Metrics} &  ~& \textbf{\mi} & \textbf{\itiu} & \textbf{\textsf{\textsc{Collab-LLM}}} &  \textbf{\textsf{Prism-SFT}}  & \textbf{\textsf{Prism-DPO}}  &~& \textbf{\textsf{LLaMA-Interact}} & \textbf{\itiu} & \textbf{\textsf{\textsc{Collab-LLM}}} &  \textbf{\textsf{Prism-SFT}}  & \textbf{\textsf{Prism-DPO}}  \\

\midrule
    ~  & \textbf{$^{\uparrow}$\textbf{Intents Cover Rate} (\%)} &  ~&   $66.84$      &     $61.92$       &    $69.78$ &    $71.05$&    $\textbf{73.62}$ &    ~&    $70.55$  &   $65.43$&     $70.27$  & $71.83$ &    $\textbf{74.85}$ \\

\rowcolor{gray!10}
~& \textbf{$^{\downarrow}$Average Interaction Turns} &~&      $3.86$         &   $\textbf{1.18}$&    $4.29$&    $1.27$&    $1.34$ & ~&        $4.21$ &   $\textbf{1.14}$&    $4.61$   &  $1.31$   & $1.39$        \\

\textbf{\inth-}& \hspace{0.5em}\textbf{Average Questions Per Turn} & ~ &    $1.10$      &    $2.42$   &   $1.73$  &   $2.36$&    $2.28$ &  ~&     $1.13$   &    $2.88$&    $1.68$  & $2.25$  & $2.21$  \\

\rowcolor{gray!10}
\textbf{\texttt{Simple}} & \textbf{$^{\uparrow}$Options Presenting Rate (\%)}&    ~&  $\textbf{85.43}$     &      $64.31$      &  $57.64$  &  $78.42$   &   $75.50$ &  ~&      $\textbf{89.32}$    &    $81.04$&    $63.10$  &   $83.61$  &  $79.12$  \\

~& \textbf{$^{\uparrow}$Options Reasonable Rate (\%)} &  ~&    $\textbf{97.28}$      &        $96.84$    &    $89.13$ &    $97.11$ &  $95.38$  &    ~&   $97.45$ &    $\textbf{98.37}$&     $90.54$ &   $98.01$   & $96.12$       \\
\rowcolor{gray!10}
~ & \hspace{0.5em}\textbf{Average Options Per Question} & ~ &      $2.59$        &   $4.02$ & $1.33$   &    $4.08$  &    $3.88$  &   ~&   $2.63$     &  $4.11$  &   $1.30$  & $4.05$  &  $3.96$       \\

\midrule

~  & \textbf{$^{\uparrow}$\textbf{Intents Cover Rate} (\%)} &  ~&    $47.81$      &     $48.93$       &    $46.35$  & $58.74$ &    $\textbf{61.02}$ & ~&    $48.72$  &   $51.34$&    $59.80$    &   $60.18$   & $\textbf{62.43}$  \\

\rowcolor{gray!10}
~& \textbf{$^{\downarrow}$Logical Conflict Rate (\%)}   &        ~&     $35.20$      &     $48.10$       &    $31.60$  &   $\textbf{11.30}$  &    $12.00$ &  ~&      $36.40$   &  $47.50$  &  $28.10$    &    $12.50$ &    $\textbf{10.80}$  \\

~& \textbf{$^{\downarrow}$Average Interaction Turns} &~&      $5.27$       &  $1.82$    &   $5.98$  &    $\textbf{1.47}$&    $1.50$  & ~&        $5.56$ &   $1.71$   &    $5.85$ &  $1.52$   & $\textbf{1.48}$       \\

\rowcolor{gray!10}
\multirow{-2}{*}{\textbf{\inth-}}& \hspace{0.5em}\textbf{Average Questions Per Turn} & ~ &     $1.28$      &       $2.72$     &    $1.83$  &    $2.96$  &    $3.10$  &  ~&      $1.22$   &   $3.22$   &    $1.54$  &   $3.41$  & $3.45$    \\

\multirow{-2}{*}{\textbf{\texttt{Complex}}} & \textbf{$^{\uparrow}$Options Presenting Rate (\%)}&    ~&  $\textbf{76.63}$     &      $56.09$      &   $44.83$ &    $72.41$  &    $73.19$ &  ~&    $72.95$   &    $72.02$&    $50.41$  &   $\textbf{73.84}$    & $71.55$   \\
\rowcolor{gray!10}
~& \textbf{$^{\uparrow}$Options Reasonable Rate (\%)} &  ~&    $73.81$      &        $67.42$   &    $74.56$ &    $90.62$  &    $\textbf{91.08}$  &    ~&   $75.12$  &  $71.03$&    $72.81$  &   $91.73$   & $\textbf{93.05}$     \\

~& \hspace{0.5em}\textbf{Average Options Per Question} & ~ &      $2.34$      &       $3.15$     &   $1.19$  &    $3.72$  &    $3.45$  &   ~&    $2.28$     &    $3.36$  &    $1.13$  &  $3.84$  & $3.78$     \\
\bottomrule
\end{tabular}
\end{center}
\vspace{-10pt}
\label{tab:clarification interaction in3}
\end{table*}

\begin{table*}[!h]
\caption{The results of clarification interaction evaluation across various metrics on the \abp test set. 
The comparisons include the \textsf{Interact} series, \itiu, \collab, and \our (with \textsf{SFT} and \textsf{DPO} variants), each utilizing two baseline model configurations.
Arrows represent the higher ($\uparrow$) or the lower ($\downarrow$) the better.
The best results are highlighted in bold.
}
\vspace{-12pt}
\begin{center}
\footnotesize
\tabcolsep=0.002\linewidth
\begin{tabular}{L{0.05\linewidth}L{0.20\linewidth}C{0.001\linewidth}|C{0.065\linewidth}C{0.06\linewidth}C{0.07\linewidth}|C{0.075\linewidth}C{0.075\linewidth}C{0.001\linewidth}|C{0.065\linewidth}C{0.06\linewidth}C{0.07\linewidth}|C{0.075\linewidth}C{0.075\linewidth}}
\toprule
\rowcolor{CadetBlue!20} 
\multicolumn{2}{c}{\textbf{\textsf{Baseline}}}&~& \multicolumn{5}{c}{\textbf{\textsf{Mistral-7B-Instruct-v0.3}}}  &   ~& \multicolumn{5}{c}{\textbf{\textsf{LLaMA-3.1-8B-Instruct}}}\\

\rowcolor{CadetBlue!20} 
\textbf{Test Data}  &\textbf{Metrics} &  ~& \textbf{\mi} & \textbf{\itiu} & \textbf{\textsf{\textsc{Collab-LLM}}} &  \textbf{\textsf{Prism-SFT}}  & \textbf{\textsf{Prism-DPO}}  &~& \textbf{\textsf{LLaMA-Interact}} & \textbf{\itiu} & \textbf{\textsf{\textsc{Collab-LLM}}} &  \textbf{\textsf{Prism-SFT}}  & \textbf{\textsf{Prism-DPO}}  \\

\midrule
    ~  & \textbf{$^{\uparrow}$\textbf{Intents Cover Rate} (\%)} &  ~&   $58.72$      &     $54.08$       &    $61.43$ &    $62.89$&    $\textbf{66.37}$ &    ~&    $63.44$  &   $58.05$&     $63.17$  & $65.02$ &    $\textbf{67.15}$ \\

\rowcolor{gray!10}
~& \textbf{$^{\downarrow}$Average Interaction Turns} &~&      $4.58$         &   $\textbf{1.43}$&    $4.95$&    $1.67$&    $1.71$ & ~&        $4.86$ &   $\textbf{1.39}$&    $5.12$   &  $1.64$   & $1.69$        \\

\textbf{\abp-}& \hspace{0.5em}\textbf{Average Questions Per Turn} & ~ &    $1.18$      &    $2.68$   &   $1.83$  &   $2.46$&    $2.31$ &  ~&     $1.21$   &    $3.01$&    $1.77$  & $2.41$  & $2.32$  \\

\rowcolor{gray!10}
\textbf{\texttt{Simple}} & \textbf{$^{\uparrow}$Options Presenting Rate (\%)}&    ~&  $\textbf{79.14}$     &      $58.23$      &  $52.84$  &  $70.67$   &   $69.11$ &  ~&      $\textbf{83.09}$    &    $74.53$&    $59.06$  &   $77.54$  &  $74.01$  \\

~& \textbf{$^{\uparrow}$Options Reasonable Rate (\%)} &  ~&    $\textbf{95.08}$      &        $94.35$    &    $87.92$ &    $95.42$ &  $93.68$  &    ~&   $96.12$ &    $\textbf{97.28}$&     $89.34$ &   $96.81$   & $94.83$       \\
\rowcolor{gray!10}
~ & \hspace{0.5em}\textbf{Average Options Per Question} & ~ &      $2.68$        &   $4.15$ & $1.26$   &    $4.09$  &    $3.82$  &   ~&   $2.71$     &  $4.22$  &   $1.25$  & $4.11$  &  $3.94$       \\

\midrule

~  & \textbf{$^{\uparrow}$\textbf{Intents Cover Rate} (\%)} &  ~&    $40.56$      &     $43.22$       &    $42.38$  & $51.43$ &    $\textbf{54.19}$ & ~&    $42.91$  &   $45.08$&    $53.01$    &   $54.77$   & $\textbf{56.12}$  \\

\rowcolor{gray!10}
~& \textbf{$^{\downarrow}$Logical Conflict Rate (\%)}   &        ~&     $44.60$      &     $56.40$       &    $38.70$  &   $18.30$  &    $\textbf{16.90}$ &  ~&      $43.80$   &  $55.60$  &  $36.10$    &    $17.10$ &    $\textbf{15.70}$  \\

~& \textbf{$^{\downarrow}$Average Interaction Turns} &~&      $5.92$       &  $2.05$    &   $6.37$  &    $\textbf{1.74}$&    $1.79$  & ~&        $6.15$ &   $1.96$   &    $6.05$ &  $1.77$   & $\textbf{1.72}$       \\

\rowcolor{gray!10}
\multirow{-2}{*}{\textbf{\abp-}}& \hspace{0.5em}\textbf{Average Questions Per Turn} & ~ &     $1.39$      &       $2.97$     &    $1.92$  &    $3.19$  &    $3.22$  &  ~&      $1.31$   &   $3.44$   &    $1.58$  &   $3.47$  & $3.49$    \\

\multirow{-2}{*}{\textbf{\texttt{Complex}}} & \textbf{$^{\uparrow}$Options Presenting Rate (\%)}&    ~&  $\textbf{70.14}$     &      $50.27$      &   $40.93$ &    $67.12$  &    $68.02$ &  ~&    $68.34$   &    $68.15$&    $46.27$  &   $\textbf{69.25}$    & $67.93$   \\
\rowcolor{gray!10}
~& \textbf{$^{\uparrow}$Options Reasonable Rate (\%)} &  ~&    $68.04$      &        $63.21$   &    $70.15$ &    $86.03$  &    $\textbf{87.44}$  &    ~&   $71.27$  &  $68.09$&    $69.38$  &   $88.61$   & $\textbf{90.34}$     \\

~& \hspace{0.5em}\textbf{Average Options Per Question} & ~ &      $2.41$      &       $3.26$     &   $1.12$  &    $3.64$  &    $3.38$  &   ~&    $2.36$     &    $3.47$  &    $1.10$  &  $3.78$  & $3.71$     \\
\bottomrule
\end{tabular}
\end{center}
\vspace{-12pt}
\label{tab:clarification interaction abp}
\end{table*}

\section{Intent Execution Experimental Details}
\label{app:Intent Execution Experimental Details}
\subsection{Metric Details}
\label{app:Metric Details2}
The metrics used for intent execution evaluation are as follows:
\begin{itemize}
\item \textbf{BLEU}: An $n$-gram-precision metric with a brevity penalty that measures overlap between the model output and the reference; higher is better.
\item \textbf{Intent Execution Faithfulness (Faithful)}: The proportion of final outputs judged to fully satisfy the clarified intent elements without omissions, distortions, or hallucinations. We utilize \textsf{GPT-4} to assist in the measurements.
\item \textbf{Unnecessary Sub-tasks (US)}: The percent of sub-tasks that are regarded as unnecessary by the user under the detailed task goal with clear user intents.
\item \textbf{General Sub-tasks (GS)}: The percent of sub-tasks that are too general, instead of focusing on the user’s specific intents.
\item \textbf{Tool Invocations Per Sub-task (TI)}: The average number of tool invocations for one sub-task, which reflects the efficiency of agent execution.
\end{itemize}

\subsection{Performance Comparison of \mi and \itiu }
\label{app:Performance Comparison of MI and ITIU}

\begin{figure}[!h]
\centering
\vspace{-7pt}
\includegraphics[width=1.0\linewidth]{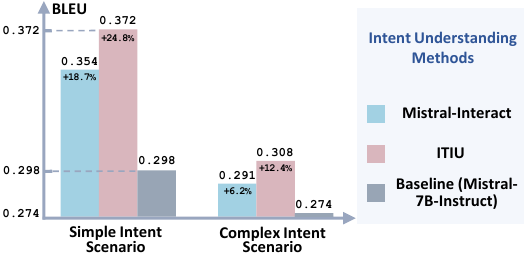}
\vspace{-17pt}
\caption{Performance comparison of \mi and \itiu across simple and complex intent scenarios.
From the simple intent scenario to the complex intent scenario, the increments of \mi and \itiu decreased from 18.8\% and 24.8\% to 6.2\% and 12.4\%
(ref. Section~\ref{sec:Evaluation on Intent Execution}).}
\Description{}
\label{fig:comparison}
  \vspace{-12pt}
\end{figure}
Figure~\ref{fig:comparison} compares the performance of \mi and \itiu under simple and complex intent scenarios.
In the simple scenario, both methods achieve substantial BLEU improvements over the baseline—\mi by 18.8\% and \itiu by 24.8\%.
However, in the complex intent scenario, their gains drop sharply to 6.2\% and 12.4\%, respectively, revealing their limited ability to model logical dependencies among clarification questions.

\section{Cognitive Load Experimental Details}
\label{app:Cognitive Load Experimental Details}
\subsection{EEG Data Preprocessing}
\label{app:EEG Data Preprocessing}

EEG signals are highly stochastic physiological signals with weak amplitudes; thus, preprocessing is essential to remove artifacts and interference before analysis. The steps are as follows:
(1) Data loading and electrode mapping: Import raw EEG data into EEGLAB and load the electrode location file (e.g., the international 10–20 system) to generate 2D or 3D topographic maps for visualization and comparison.
(2) Filtering: Apply a 0.1 Hz high-pass and 40 Hz low-pass (or 0.1–40 Hz band-pass) filter, along with a 48–52 Hz notch filter to eliminate 50 Hz power line noise.
(3) Artifact removal via ICA: Perform Independent Component Analysis (ICA) to separate signal sources, identify and remove artifacts (e.g., ocular, cardiac, and muscle activity), and reconstruct the cleaned EEG signals.
(4) Re-referencing: Compute the average signal across all channels as a reference and subtract it from each electrode’s potential to reduce recording bias.
(5) Segmentation: Divide the continuous EEG data into segments using a 10 s sliding window with a 1 s step size for subsequent analysis.

\subsection{Frequency Domain Transformation}
\label{app:Frequency Domain Transformation}
We convert the time-domain EEG signals into frequency-domain signals for subsequent analysis using the Fast Fourier Transform (FFT) algorithm.

Assuming the EEG signal sampling rate is $F_s$, and the number of samples in an EEG segment is $N$, with zeros padded at the end of the data to extend it to the nearest power of 2, the frequency domain is discretized as $f=kF_s/N,k=0,1,...,N-1$. The frequency-domain representation of the signal after transformation is given by:

\begin{equation}F[k]=\sum_{n=0}^{N-1}x[n]e^{-j2\pi kn/N}.\end{equation}

Based on the Fourier Transform, we can calculate the Power Spectral Density (PSD) of the signal, which describes how the signal's power varies with frequency. 
The unit of PSD is $\mathrm{V}^{2}/\mathrm{Hz}$ or $dB$, i.e., $10log_{10}(\mathrm{V}^2/\mathrm{Hz})$. 
We calculate the EEG signal's PSD using the periodogram method:

\begin{equation}P(f)=\frac{1}{NF_s}\left|\sum_{n=0}^{N-1}x[n]w[n]e^{-j2\pi fn/F_s}\right|,\end{equation}
where $w[n]$ is a window function, $w[n]=0.5(1-\cos(\frac{2\pi n}{N-1}))$, used to assign different weights to the signal samples.

The PSD in various EEG frequency bands $\delta\mathrm{(1-3Hz)}$, $\theta\mathrm{~(3-8Hz)}$, $\alpha\mathrm{~(8-13Hz)}$, and $\beta(13-30Hz)$ can be calculated as:

\begin{equation}P(band)=\frac{1}{N_{band}}\sum_{f=band_{min}}^{band_{max}}P(f),band=\delta,\theta,\alpha,\beta,\end{equation}
where $N_{band}$ is the number of frequency values within the transformed frequency band, and $band_{min}$ and $band_{max}$ define the range of the band in the frequency domain.

% We calculate the PSD of EEG data from all trials across different frequency bands and average it across channels to reveal the overall trend of the power spectral density changes in the cognitive load experiment.
% During the experiment, the energy in the high cognitive load state is significantly higher than in the low load state. The mean and variance of the EEG power spectral density are both larger, and the changes in power spectral energy are more pronounced.

We computed the PSD of EEG data from all trials across different frequency bands and averaged it across channels to examine overall trends in spectral power during the cognitive load experiment.
The results show that EEG energy in the high cognitive load state is significantly greater than in the low load state, with higher mean and variance of PSD values and more pronounced fluctuations in spectral power.

% \section{Experiment on Self-evolved Rounds}

\end{document}